\documentclass{article}
\usepackage{guidance_preprint}

\usepackage{amsmath,amsfonts,bm}

\def\eqref#1{equation~\ref{#1}}

\def\1{\bm{1}}

\DeclareMathAlphabet{\mathsfit}{\encodingdefault}{\sfdefault}{m}{sl}
\SetMathAlphabet{\mathsfit}{bold}{\encodingdefault}{\sfdefault}{bx}{n}

\usepackage[utf8]{inputenc}
\usepackage[T1]{fontenc}
\usepackage{url}
\usepackage{graphicx}
\usepackage{booktabs}
\usepackage{amsmath}
\usepackage{amssymb}
\usepackage{microtype}
\usepackage{array}
\usepackage{tabularx}
\usepackage{makecell}
\usepackage{float}
\usepackage{placeins}
\usepackage{needspace}
\usepackage{flafter}
\usepackage[table]{xcolor}
\usepackage{fontawesome5}
\usepackage{booktabs}
\usepackage{multirow}
\usepackage{threeparttable}
\usepackage{algorithm}
\usepackage{algpseudocode}
\usepackage{listings}
\usepackage[most]{tcolorbox}
\usepackage{tikz}
\usetikzlibrary{arrows.meta,calc,fit,positioning}
\usetikzlibrary{shadows}
\usepackage{hyperref}
\hypersetup{hidelinks}

\providecommand{\nolinenumbers}{}
\providecommand{\linenumbers}{}
\definecolor{methodblue}{HTML}{24527A}
\definecolor{methodblueback}{HTML}{F2F7FB}
\definecolor{trainorange}{HTML}{C75B12}
\definecolor{trainorangeback}{HTML}{FFF5EC}
\definecolor{verifygreen}{HTML}{277A4B}
\definecolor{verifygreenback}{HTML}{EFF8F2}

\newtcolorbox{takeaway}[1]{
    enhanced,
    colback=methodblueback,
    colframe=methodblue,
    colbacktitle=methodblue,
    coltitle=white,
    fonttitle=\bfseries\small,
    fontupper=\small,
    title={#1},
    boxrule=0.6pt,
    arc=1.5mm,
    left=2mm,
    right=2mm,
    top=1.3mm,
    bottom=1.3mm,
    before skip=6pt,
    after skip=7pt
}

\title{Evolving in Thought Space: Training a Small Model\\
at Test Time Unlocks Better Discoveries}

\author{%
\parbox{\dimexpr\textwidth-2\tabcolsep\relax}{\centering
\normalfont\normalsize
Chonghe Jiang\textsuperscript{1,9,*}\quad
Ao Qu\textsuperscript{1,9,*}\\[4pt]
Siyuan Liu\textsuperscript{2}\quad
Ruoyun Ma\textsuperscript{3,\ensuremath{\ddagger}}\quad
Zijian Zhou\textsuperscript{4}\quad
Dingyi Zhuang\textsuperscript{1}\\[4pt]
Bo Liu\textsuperscript{5,6}\quad
Han Zheng\textsuperscript{1}\quad
Hanfei Yu\textsuperscript{7}\quad
Baichuan Mo\textsuperscript{8}\\[4pt]
Jinhua Zhao\textsuperscript{1,9,\ensuremath{\dagger}}\quad
Paul Pu Liang\textsuperscript{1,9,\ensuremath{\dagger}}\\[8pt]
\normalfont\fontsize{8.5}{10}\selectfont
\textsuperscript{1}\,Massachusetts Institute of Technology\quad
\textsuperscript{2}\,Hong Kong Polytechnic University\quad
\textsuperscript{3}\,ByteDance Inc.\\
\textsuperscript{4}\,National University of Singapore\quad
\textsuperscript{5}\,Stanford University\quad
\textsuperscript{6}\,University of Washington\\
\textsuperscript{7}\,University of California, Berkeley\quad
\textsuperscript{8}\,Tsinghua University\\
\textsuperscript{9}\,Singapore-MIT Alliance for Research and Technology\\[4pt]
\textsuperscript{*}\,Equal contribution.\quad
\textsuperscript{\ensuremath{\dagger}}\,Joint advising.\\
\textsuperscript{\ensuremath{\ddagger}}\,Work done outside the authors' employment.
}}
\date{}

\hypersetup{
  pdftitle={Evolving in Thought Space: Training a Small Model at Test Time Unlocks Better Discoveries},
  pdfauthor={Chonghe Jiang, Ao Qu, Siyuan Liu, Ruoyun Ma, Zijian Zhou, Dingyi Zhuang, Bo Liu, Han Zheng, Hanfei Yu, Baichuan Mo, Jinhua Zhao, Paul Pu Liang},
  pdfsubject={Guidance-TTT}
}

\begin{document}

\maketitle

\begin{abstract}
Open-ended scientific discovery often requires repeatedly proposing and
evaluating candidate solutions. LLM-based systems can support this process by
generating and refining executable solutions from verifier feedback. Methods
such as TTT-Discover use test-time training (TTT) to update the
solution-generating LLM from verifier feedback, adapting its generation policy
to improve subsequent proposals on the target problem. However, this becomes expensive when reliable execution requires a large
model, since training must maintain gradients, optimizer states, and policy
statistics while repeatedly generating long, structured outputs. It also
complicates credit assignment: outcome-level verifier feedback must jointly
evaluate the high-level strategy and its low-level implementation. In this work, we introduce \emph{Guidance-TTT}, which separates these roles. A compact
\emph{guidance model} is trained at test time to propose high-level strategic
changes, while a frozen \emph{execution model} implements them as
complete executable solutions. At each step, the system selects a promising
previously discovered solution, proposes a change, executes and verifies it,
and updates only the guidance model using an adaptive group-relative RL
objective. This concentrates test-time learning on short strategic decisions
while retaining the implementation capability of a substantially stronger
model without adapting it. Without web access, Guidance-TTT produces strong solutions across four distinct domains: combinatorial optimization (Polyomino Packing), heuristic programming (AHC058), machine learning (Lasso), and GPU kernel optimization (TriMul). Across these tasks, it outperforms the best solutions reported in prior work while remaining competitive with state-of-the-art results on public online leaderboards. Code is available at
\url{https://github.com/Human-Agent-Society/reef/tree/main/recipes/tttd/examples/guidance_ttt}.
\end{abstract}

\begingroup
\setlength{\intextsep}{6pt}
\begin{table}[H]
\centering
\begin{minipage}{0.94\linewidth}
\centering
\begin{threeparttable}
\caption{Discovery results across four domains.}
\label{tab:discovery-highlights}
\footnotesize
\setlength{\tabcolsep}{4pt}
\renewcommand{\arraystretch}{1.08}
\begin{tabularx}{\linewidth}{@{}l*{4}{>{\centering\arraybackslash}X}@{}}
\toprule
& \makecell{\textbf{Polyomino}\\\citep{qu2026coral}}
& \makecell{\textbf{Lasso}\\\citep{ye2026evaluation}}
& \makecell{\textbf{AHC058}\\\citep{ye2026evaluation}}
& \makecell{\textbf{TriMul}\stepcounter{footnote}\hyperlink{opening.trimul}{\textsuperscript{\thefootnote}}\\\citep{cao2026ksearch}} \\
Method & Score $\uparrow$
& Score $\uparrow$
& AtCoder score $\uparrow$
& Latency ($\mu$s) $\downarrow$ \\
\midrule
Previous SOTA\stepcounter{footnote}\hyperlink{opening.sota}{\textsuperscript{\thefootnote}} & 89.40 & 0.1243 & 849,325,750 & 1131 \\
Ours & \textbf{91.89} & \textbf{0.1739} & \textbf{850,082,731} & \textbf{1129} \\
\bottomrule
\end{tabularx}
\end{threeparttable}
\end{minipage}
\end{table}
\endgroup
\makeatletter
\begingroup
\def\Hy@footnote@currentHref{opening.trimul}
\addtocounter{footnote}{-1}
\footnotetext{TriMul latencies use an H100 80GB GPU with CUDA 12.8, PyTorch 2.7.1, and Triton 3.3.1. See Section~\ref{sec:trimul-setup} and Appendix~\ref{app:fixed-evaluation} for evaluation details.}
\def\Hy@footnote@currentHref{opening.sota}
\addtocounter{footnote}{1}
\footnotetext{“Previous SOTA” refers to the best solutions reported in prior work, rather than the current best entries on public leaderboards.}
\endgroup
\makeatother


\section{Introduction}
\label{sec:introduction}
Scientific discovery can be enhanced by LLMs that iteratively
propose, evaluate, and refine candidate solutions
\citep{romeraparedes2024funsearch,novikov2025alphaevolve,qu2026coral}.
Many existing systems keep the LLM frozen and use verifier feedback to select
promising candidates and condition subsequent generations through the prompt
\citep{romeraparedes2024funsearch,novikov2025alphaevolve}.
This supports inference-time search and in-context adaptation without changing
model parameters. Test-time training (TTT) additionally updates model parameters
using verifier feedback to adapt the generation policy to the current problem
\citep{wang2025thetaevolve,yuksekgonul2026learning}.

Existing discovery-oriented TTT methods train the same LLM that
generates the complete candidate solution
\citep{surina2025algorithm,wang2025thetaevolve,yuksekgonul2026learning}.
This coupling creates two challenges.
First, reliable solution generation often requires a large model, making
direct test-time adaptation expensive in memory and computation
\citep{hu2022lora,yuksekgonul2026learning}.
Second, verifier feedback is applied to the complete output, so a single
outcome-level reward must assign credit across both high-level strategy and
low-level implementation
\citep{setlur2025rewarding,samanta2026credit}.
Consequently, a promising strategy may receive poor feedback simply because
its implementation fails. These limitations motivate separating strategic guidance from solution
implementation: test-time adaptation focuses on deciding what to change, while
a capable executor handles how to realize it. This division also reflects how
human experts often solve difficult problems through a small number of critical
strategic decisions rather than by reconsidering every implementation detail.

\begin{figure}[!htb]
\centering
\includegraphics[width=\linewidth]{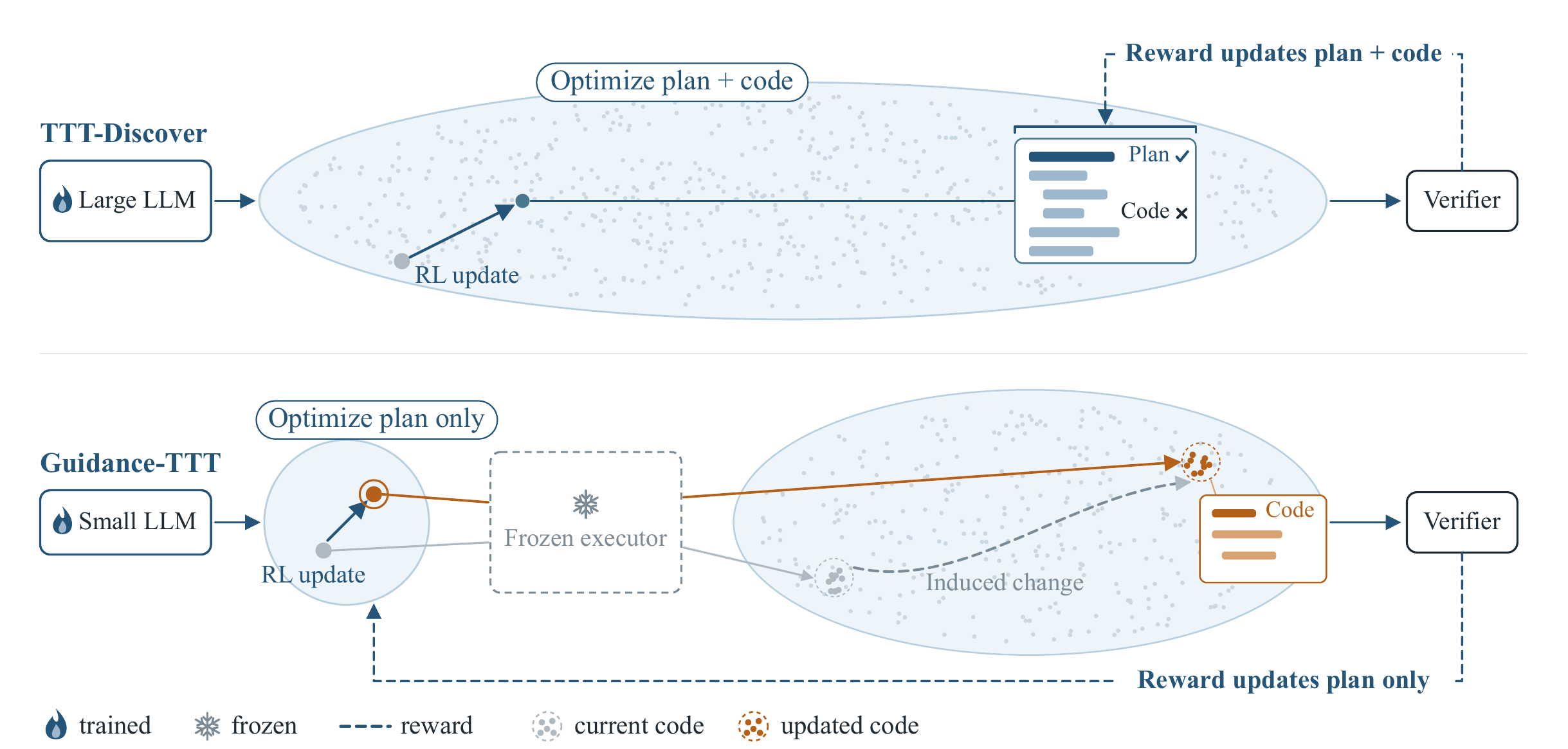}
\caption{\textbf{Test-time learning in plan space.}
TTT-Discover uses verifier reward to update both plan and code generation.
Guidance-TTT applies the reward only to the small guidance model, while a
frozen executor translates plan changes into code changes.}
\label{fig:guidance-plan-space}
\end{figure}

\begin{figure}[!t]
\centering
\includegraphics[width=\linewidth,trim=0 5bp 0 5bp,clip]{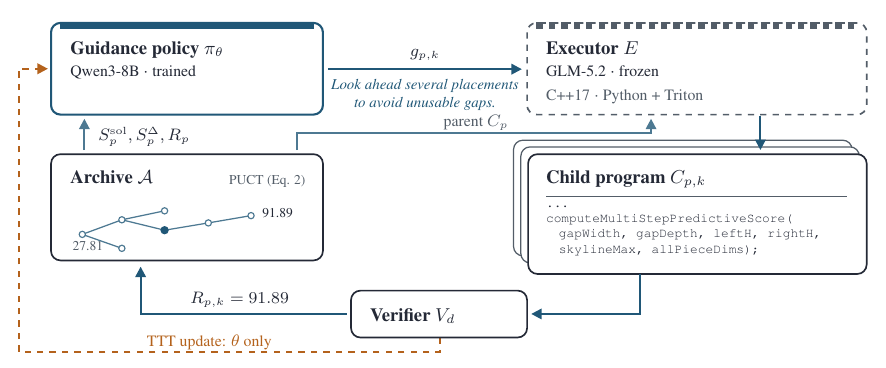}
\caption{Guidance-TTT workflow. A PUCT-based archive search selects a promising parent solution, the guidance
model proposes a high-level modification, and the frozen executor realizes it as
a new candidate. The resulting verifier reward updates only the guidance model.}
\label{fig:hierarchical-ttt-overview}
\end{figure}

We therefore introduce \emph{Guidance-TTT}
(Figure~\ref{fig:guidance-plan-space}), with the full workflow shown in
Figure~\ref{fig:hierarchical-ttt-overview}. A compact guidance model is trained
at test time to propose high-level, parent-conditioned changes, while a stronger
execution model remains frozen and implements them as complete candidate
solutions. Only the guidance model is updated, reducing adaptation cost and
focusing test-time learning on strategic decisions while preserving the
implementation capability of the frozen executor. We refer to this as learning in \emph{thought space}, where the learned actions
are explicit natural-language proposals rather than complete programs or latent
representations. After each execution, the system stores the resulting candidate
together with compact summaries describing the solution and its change from the
parent. The guidance model conditions on these summaries and the verifier score,
while the execution model receives the exact parent solution together with the
proposed guidance.

Across four tasks, we train only 8B or 14B guidance while keeping GLM-5.2
frozen, without web access (Table~\ref{tab:main-results}). Polyomino reaches
91.89, exceeding four-agent, web-enabled CORAL's 89.40 by 2.49 points.
Our Lasso solver is approximately \(1.40\times\) faster than SimpleTES under
the same correctness tolerance. On AHC058, we obtain 850.08M AtCoder points
versus SimpleTES's 849.33M. On TriMul, we reach \(1129\,\mu\text{s}\), close
to the public rank-1 reference of \(1128\,\mu\text{s}\).
In a separate Polyomino comparison using GPT-OSS-120B for execution,
adapting only 8B guidance costs about 58\% less than directly adapting the
120B model, with comparable scores
of 83.19 and 83.72, respectively (Section~\ref{sec:ablations}).

\begin{samepage}
We make two contributions:
\begin{enumerate}
    \item We introduce \textbf{Guidance-TTT}, a test-time RL framework that
    decouples strategic adaptation from solution generation: a compact guidance
    model learns parent-conditioned changes in thought space, while a stronger
    execution model remains frozen. We demonstrate its performance across four modern tasks ranging from algorithm design to kernel design.

    \item We analyze how guidance learning affects discovery through controlled
    ablations, model-scale comparisons, search trajectories, and failure-mode
    studies, revealing  the benefits of learning in thought
    space.
\end{enumerate}
\end{samepage}

\begin{figure}[!tb]
\centering
\includegraphics[width=\linewidth]{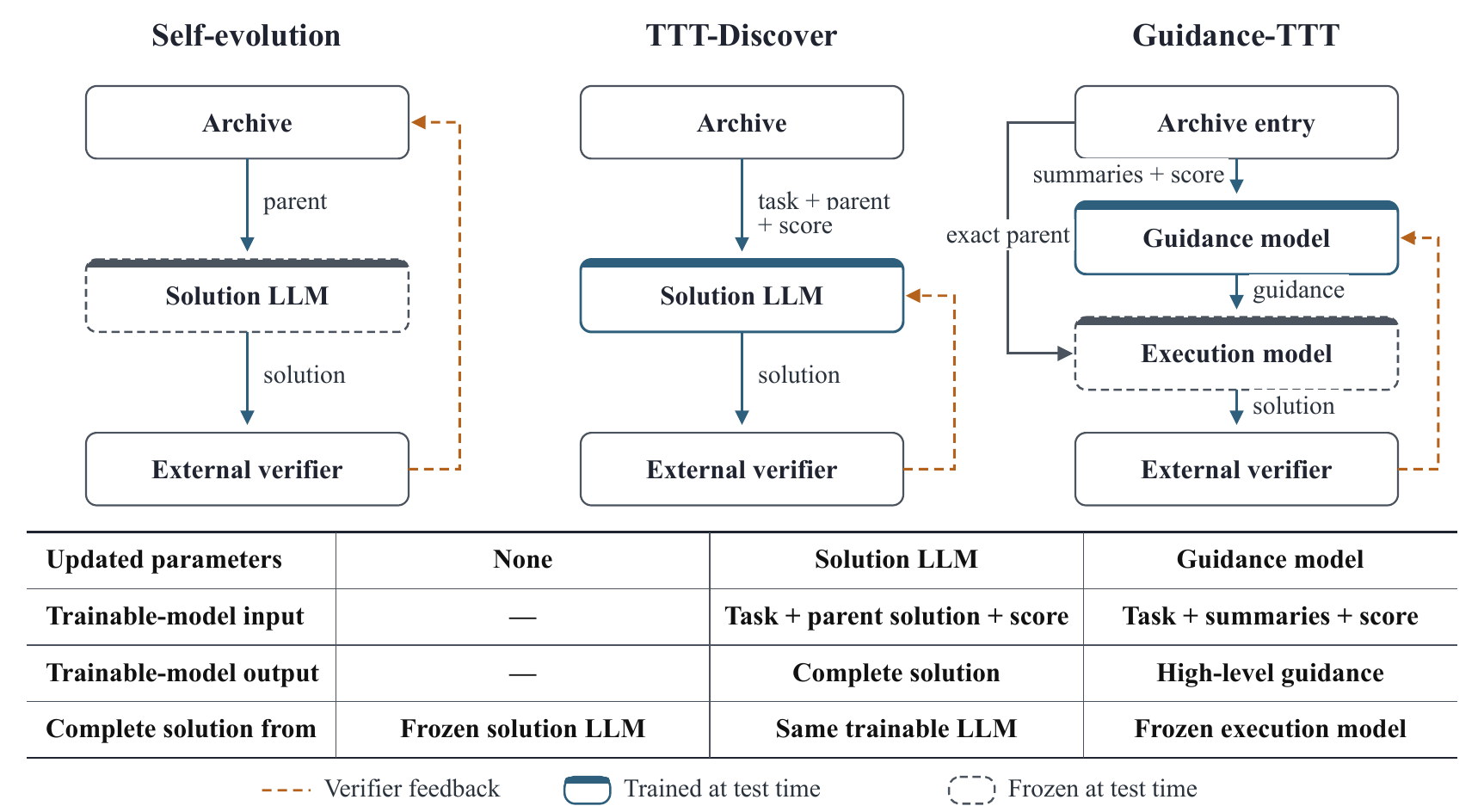}
\caption{Discovery paradigms differ in where verifier feedback drives adaptation: search in self-evolution, trainable solution generation in TTT-Discover, and trainable strategic guidance in Guidance-TTT. Training only the compact guidance model reduces adaptation cost while retaining the implementation capabilities of a stronger, frozen executor.}
\label{fig:method-comparison}
\end{figure}

\section{Related Work}
\label{sec:related-work}

\noindent\textbf{LLM self-evolution.}
A growing line of work improves LLM-generated solutions through iterative
inference-time search without updating model parameters. FunSearch alternates
program generation, selection, and evaluation to discover mathematical
constructions and heuristics \citep{romeraparedes2024funsearch}. AlphaEvolve
extends this paradigm with a general-purpose coding model and an evolutionary
program database \citep{novikov2025alphaevolve}. CORAL further distributes
open-ended discovery across multiple fixed frontier LLMs using shared memory
and asynchronous evolution \citep{qu2026coral}. These methods demonstrate that
archive reuse, verifier feedback, and repeated refinement can drive substantial
improvement while keeping the underlying models fixed.

\noindent\textbf{Test-time training.}
Classical test-time training adapts a model using a learning problem constructed
from each test instance \citep{sun2020ttt,sun2023learning}. Recent LLM methods
extend this idea to reasoning and open-ended discovery by updating model
parameters during inference. ThetaEvolve augments a
program database with optional test-time RL \citep{wang2025thetaevolve} and
TTT-Discover introduces an entropic objective with PUCT-based reuse for
max-seeking discovery \citep{yuksekgonul2026learning}. 

\noindent\textbf{Role specialization in LLM systems.}
Prior work has explored decomposing LLM systems into specialized roles that
coordinate through natural-language communication \citep{wu2023autogen}.
More recent work explicitly separates ideation from execution.
\citet{si2026execution} develop an automated executor that implements
LLM-generated research ideas and evaluates them through large-scale experiments,
allowing the ideation process to be grounded in execution feedback.
Guidance-TTT learns parent-conditioned guidance from verifier feedback at
test time, rather than only assigning ideation and execution to separate models.

\section{Guidance-TTT}
\label{sec:method}

\subsection{Problem Setup}
\label{sec:preliminaries}

Following TTT-Discover \citep{yuksekgonul2026learning}, we consider a fixed
discovery problem specified by a task description \(d\) and an executable
verifier \(V_d\). Each candidate \(C\in\mathcal{C}_d\) is a complete solution,
and its reward is \(R=V_d(C)\). In Polyomino Packing, for example, the task
description specifies the input and validity constraints, the candidate is a
complete C++ solver, and the verifier compiles and evaluates it on the benchmark
instances.

Let \(\mathcal{A}_u\) denote the archive after update \(u\). At each step,
the system refines a previously evaluated solution to produce a new candidate,
which is evaluated by the verifier and may be added to the archive. The goal is
to discover the highest-reward solution by the end,
\[
C^\star
=
\arg\max_{(C_i,R_i)\in\mathcal{A}_U} R_i.
\]

\subsection{What Should Be Learned at Test Time?}

LLM-based methods for scientific discovery differ in what they learn from verifier feedback.
In self-evolution without parameter updates, a fixed solution LLM generates the
complete child from the task and its selected parent. Feedback changes which
solutions are retained or selected for later refinement, but it does not change
the model \citep{romeraparedes2024funsearch,novikov2025alphaevolve}.

TTT-Discover and related discovery-oriented methods
\citep{surina2025algorithm,wang2025thetaevolve,yuksekgonul2026learning}
instead use evaluated rollouts to update the same LLM that generates complete
candidate solutions. The model must decide what change to attempt and implement
that change in a complete solution. This is expensive when reliable generation
requires a large model, and outcome-level feedback combines the quality of the
strategy with the quality of its implementation.

These limitations motivate separating strategic adaptation from solution
implementation. Guidance-TTT therefore updates only a compact guidance model
that proposes high-level changes, while a separate execution model remains
frozen and applies them to the exact parent solution. This focuses test-time
learning on strategic decisions while avoiding the cost of adapting a much
larger solution generator. Figure~\ref{fig:method-comparison} summarizes the
distinction from self-evolution and TTT-Discover.

\subsection{The Guidance-TTT Loop}
\label{sec:hierarchical-interface}
We next describe the Guidance-TTT loop, including parent selection, guidance
generation, execution, verification, and the test-time update. Algorithm~\ref{alg:hierarchical-ttt} summarizes the Guidance-TTT loop, and
Figure~\ref{fig:hierarchical-ttt-overview} shows the information exchanged in
one iteration.

\begin{algorithm}[t]
\caption{Guidance-TTT for one test-time run on problem \(d\)}
\label{alg:hierarchical-ttt}
\begin{algorithmic}[1]

\Require task \(d\), verifier \(V_d\), seed solution \(C_0\)
\Require frozen execution model \(E\), guidance policy \(\pi_\theta\)
\Require updates \(U\), parent groups per update \(B\), rollouts per group \(K\)

\State \(R_0 \gets V_d(C_0)\)
\State initialize \(S^{\mathrm{sol}}_0\) from \(C_0\) and set
\(S^{\Delta}_0 \gets \varnothing\)
\State \(\mathcal{A} \gets
\{(C_0,S^{\mathrm{sol}}_0,S^{\Delta}_0,R_0)\}\)
\Comment{parent-linked archive}

\For{\(u=1,\ldots,U\)}

    \State \((p_1,\ldots,p_B) \gets
    \operatorname{PUCTSelect}(\mathcal{A},B)\)
    \Comment{repeats allowed; Appendix~\ref{app:method-details}}

    \For{\(b=1,\ldots,B\), with parent \(p=p_b\)}
        \For{\(k=1,\ldots,K\)}

            \State \(g_{b,k}\sim
            \pi_\theta\!\left(
            \cdot\mid
            d,S^{\mathrm{sol}}_p,S^{\Delta}_p,R_p
            \right)\)
            \Comment{propose guidance}

            \State \((C_{b,k},
            S^{\mathrm{sol}}_{b,k},
            S^{\Delta}_{b,k})
            \gets E(d,C_p,g_{b,k})\)
            \Comment{modify exact parent}

            \State \(R_{b,k}\gets V_d(C_{b,k})\); record the attempt and feedback

            \State add \((C_{b,k},S^{\mathrm{sol}}_{b,k},S^{\Delta}_{b,k},R_{b,k})\)
            to \(\mathcal{A}\) only if archive-admissible, linked to \(p\)

        \EndFor
    \EndFor

    \State \(\theta \gets
    \operatorname{GuidanceUpdate}
    \bigl(
        \theta,
        \{(p_b,g_{b,k},R_{b,k})\}_{b=1,\,k=1}^{B,\,K}
    \bigr)\)
    \Comment{group-relative TTT update}
    \State apply archive retention rules \Comment{Appendix~\ref{app:method-details}}

\EndFor

\State \Return
\(\displaystyle
\arg\max_{(C,\cdot,\cdot,R)\in\mathcal{A}} R\)

\end{algorithmic}
\end{algorithm}

Each eligible solution is stored in the archive together with compact summaries
and its verifier score. Given a selected parent, the guidance model conditions
on its summaries and score to sample \(K\) high-level modifications. The frozen
execution model applies each modification to the exact parent solution, producing
\(K\) child candidates that are evaluated by the verifier. Their rewards form a
sibling group used to update only the guidance policy.

We select parents from the solution archive using PUCT sampling rule
\citep{yuksekgonul2026learning}. For a node \(i\), we define its value by the
best oriented task score among its direct children,
\begin{equation}
Q(i)
=
\begin{cases}
v(i), & \mathcal{C}(i)=\varnothing,\\[1mm]
\displaystyle\max_{j\in\mathcal{C}(i)}v(j),
& \text{otherwise},
\end{cases}
\end{equation}
and select parents according to
\begin{equation}
\operatorname{PUCT}(i)
=
Q(i)
+
c\,s\,P(i)
\frac{\sqrt{1+N_{\mathrm{exp}}}}{1+n_i},
\label{eq:puct}
\end{equation}
where \(n_i\) is the number of times node \(i\) has been expanded,
\(N_{\mathrm{exp}}\) is the total number of expansions, \(P(i)\) is a
rank-based prior, and \(s\) normalizes the exploration term by the current
score range. The oriented score \(v\) and RL reward mappings are specified
in Appendix~\ref{app:method-details}.

\subsection{Guidance Model Update via RL}
\label{sec:search-training}

We train the guidance policy to propose changes that yield high-reward
solutions when implemented by the frozen executor. 
For a selected parent \(p\) with compact state \(s_p\), we maximize the
entropic reward objective \citep{yuksekgonul2026learning}
\begin{equation}
J_{\beta}(\theta\mid p)
=
\frac{1}{\beta}\log
\mathbb{E}_{g \sim \pi_\theta(\cdot \mid s_p)}
\mathbb{E}_{C \sim E(\cdot \mid C_p, g)}
\left[\exp\!\left(\beta V_d(C)\right)\right],
\label{eq:guidance-rl}
\end{equation}
where \(\pi_\theta\) is the guidance policy, \(E\) is the frozen executor,
and \(\beta>0\) controls the emphasis on high rewards. The exponential
weighting favors high-reward candidates over simply improving mean reward.

For each parent, we sample \(K\) guidance actions and independently evaluate
their realized solutions. The resulting rewards form a group-relative learning
signal \citep{shao2024deepseekmath}. We adapt \(\beta\) within each group and
hold it fixed during the policy update, treating the selected parent as a fixed
training context. The guidance policy is then optimized using an entropic
advantage estimator with a centered base-policy correction
\citep{yuksekgonul2026learning}. Appendix~\ref{app:rl-update} gives the full
estimator, correction, and actor loss.

\section{Experiments}
\label{sec:experiments}
We evaluate Guidance-TTT on four discovery tasks: Polyomino Packing,
TriMul GPU kernel optimization, Lasso path optimization, and AHC058.
Unless otherwise stated, we train Qwen3-8B
\citep{yang2025qwen3} with rank-32 LoRA \citep{hu2022lora} as the guidance
model and keep GLM-5.2 \citep{glm5team2026glm5} frozen as the execution
model. Each run uses 30 updates, with \(B=8\) selected parents and \(K=16\)
rollouts per parent. Full settings are provided in Appendix~\ref{app:settings}.
\subsection{Experimental Setup}
Seeds are generated by one-shot execution-model calls; discovery is entirely
offline, without web access or inheriting existing solution code.

\phantomsection
\noindent\textbf{Polyomino Packing.}
\label{sec:polyomino}
\label{sec:polyomino-setup}
We use FrontierCS problem 0 \citep{mang2025frontiercs}, where each candidate is
a C++17 solver for packing irregular pieces across 70 cases. Performance is
measured by aggregated packed-cell density on a \(0\)--\(100\) scale, with
higher being better.

\phantomsection
\noindent\textbf{TriMul GPU kernel.}
\label{sec:trimul}
\label{sec:trimul-setup}
TriMul is the outgoing Triangle Multiplicative Update kernel from AlphaFold3
\citep{abramson2024alphafold3}, following the TTT-Discover GPU benchmark
\citep{yuksekgonul2026learning}. Candidates are Triton-based Python kernels that
must pass correctness tests before timing. We report geometric-mean latency
over seven benchmark cases using repeated fixed-kernel evaluations, with lower
being better. Public ranks refer to the cited GPUMode leaderboard snapshot
\citep{gpumodetrimul}. We include ranks 1, 4, and 5; ranks 2 and 3 are excluded
because they use compiled CUDA/C++ extensions rather than Triton kernels.
Details are in Appendix~\ref{app:trimul-compiler-transfer}.

\noindent\textbf{Lasso regularization path.}
Candidates are double-precision C++ solvers for the full Gaussian Lasso path.
Valid solutions must match the scikit-learn reference objective within
\(10^{-6}\). We evaluate fixed solvers repeatedly and report the reciprocal of
geometric-mean solve time, with higher being better. The TTT-Discover Lasso
reference is from our own evaluation.

\noindent\textbf{AHC058: Apple Production Planning.}
AHC058 is a 500-turn planning task in which the solver allocates resources
across four machine levels to maximize apple production. Performance is
measured by the task score, with higher being better; final results use the
submitted solution's AtCoder score.

\noindent\textbf{Comparison methods.}
We compare with TTT-Discover, CORAL, OpenEvolve, and task-specific references.
The verified TTT-Discover runs use GPT-OSS-120B. The GLM-5.2 CORAL
Polyomino run uses a 12\,h budget, and OpenEvolve uses GLM-5.2 for 500 iterations.
For Polyomino, we also include CORAL's offline Opus 4.6 result and the
FrontierCS human best. The opening comparison in
Table~\ref{tab:discovery-highlights} instead uses CORAL's stronger four-agent,
web-enabled result (89.40).
Complete evaluation protocols are in Appendix~\ref{app:fixed-evaluation}.

\subsection{Main Results}
Table~\ref{tab:main-results} summarizes results from prior work, public references and our runs. We report the model configuration associated with
each result. For Guidance-TTT, the two models correspond to the guidance and
execution models, respectively. N/A indicates either a human reference or a
public result whose model configuration could not be reliably verified. We
discuss the resulting performance and discovered solutions below, with
controlled ablations presented in Section~\ref{sec:ablations}. Detailed configuration can be found in
Appendix~\ref{app:fixed-evaluation}.

\begin{table}[!t]
\centering
\begin{threeparttable}
\caption{Comparison with prior work and public references across four discovery tasks.
Bold marks the best displayed value per task.}
\label{tab:main-results}
\small
\setlength{\tabcolsep}{4pt}
\renewcommand{\arraystretch}{1.08}
\begin{tabularx}{\linewidth}{@{}p{0.46\linewidth}Xr@{}}
\toprule
Method / solution & Model(s) & Result \\
\midrule
\multicolumn{3}{@{}l}{\textbf{Polyomino Packing} \quad Score $\uparrow$; 70 cases} \\
CORAL \citep{qu2026coral} & Claude Opus 4.6 & 84.20 \\
CORAL & GLM-5.2 & 83.80 \\
TTT-Discover \citep{yuksekgonul2026learning} & GPT-OSS-120B & 83.72 \\
OpenEvolve \citep{sharma2025openevolve} & GLM-5.2 & 78.36 \\
Human best \citep{mang2025frontiercs} & N/A & 89.10 \\
\rowcolor[gray]{0.93}
Ours & Qwen3-8B + GLM-5.2 & \textbf{91.89} \\
\midrule
\multicolumn{3}{@{}l}{\textbf{Lasso Path} \quad Score $\uparrow$; 17 cases} \\
SimpleTES \citep{ye2026evaluation} & GPT-OSS-120B & 0.1243 \\
TTT-Discover \citep{yuksekgonul2026learning} & GPT-OSS-120B & 0.1368 \\
OpenEvolve \citep{sharma2025openevolve} & GLM-5.2 & 0.1402 \\
\rowcolor[gray]{0.93}
Ours & Qwen3-8B + GLM-5.2 & \textbf{0.1739} \\
\midrule
\multicolumn{3}{@{}l}{\textbf{AHC058} \quad AtCoder score $\uparrow$} \\
TTT-Discover \citep{yuksekgonul2026learning} & GPT-OSS-120B & 848,414,228 \\
SimpleTES \citep{ye2026evaluation} & GPT-OSS-20B & 849,325,750 \\
OpenEvolve \citep{sharma2025openevolve} & GLM-5.2 & 809,885,724 \\
\rowcolor[gray]{0.93}
Ours & Qwen3-14B + GLM-5.2 & \textbf{850,082,731} \\
\midrule
\multicolumn{3}{@{}l}{\textbf{TriMul} \quad Geometric-mean latency ($\mu$s) $\downarrow$; H100} \\
GPUMode rank 1 \citep{gpumodetrimul} & N/A & \textbf{1128} \\
GPUMode rank 4 & N/A & 1163 \\
GPUMode rank 5 & N/A & 1278 \\
K-Search \citep{cao2026ksearch} & GPT-5.2 + Gemini-3-Pro & 1131 \\
TTT-Discover \citep{yuksekgonul2026learning} & GPT-OSS-120B & 1262\tnote{a} \\
CORAL \citep{qu2026coral} & GLM-5.2 & 1180 \\
OpenEvolve \citep{sharma2025openevolve} & GLM-5.2 & 2152 \\
\rowcolor[gray]{0.93}
Ours & Qwen3-14B + GLM-5.2 & 1129 \\
\bottomrule
\end{tabularx}
\begin{tablenotes}[flushleft]\footnotesize
\item[a] Re-evaluated in our environment; the original paper reports 1161\,$\mu$s.
See Appendix~\ref{app:trimul-compiler-transfer}.
\end{tablenotes}
\end{threeparttable}
\end{table}

\subsubsection{Performance Analysis}
\label{sec:polyomino-main-performance}
\label{sec:trimul-main-performance}
Guidance-TTT achieves stronger final solutions than the prior-work references
across all four tasks, without web access or adapting the execution model.
The gains are particularly pronounced on Polyomino and Lasso: Polyomino
reaches 91.89, surpassing both web-enabled four-agent CORAL (89.40) and the
released human reference (89.10), while Lasso improves over SimpleTES from
0.1243 to 0.1739, a 39.8\% gain under the same correctness tolerance. On the
already strong AHC058 and TriMul benchmarks, our official AHC058 submission
reaches 850.08M points, exceeding SimpleTES (849.33M) and TTT-Discover
(848.41M), while our TriMul kernel achieves 1129\,$\mu$s geometric-mean latency,
slightly outperforming K-Search (1131\,$\mu$s) and approaching the eligible
public rank-1 result (1128\,$\mu$s). 

Appendix~\ref{app:further-results} tests whether these gains persist beyond
the primary evaluation settings. On Lasso, the solvers discovered by
Guidance-TTT with GLM-5.2 execution pass all 15 held-out real and stress-test cases, compared with 13/15
for SimpleTES. On TriMul, the discovered kernel transfers unchanged across
Triton versions and remains the fastest evaluated automated method on Triton
3.4.0 and 3.6.0. We additionally report alternative execution-model results
for Polyomino.
\subsubsection{Solution Analysis}
\label{sec:solution-analysis}
\begin{figure}[!htbp]
\centering
\includegraphics[width=\linewidth,trim=0 6bp 0 6bp,clip]{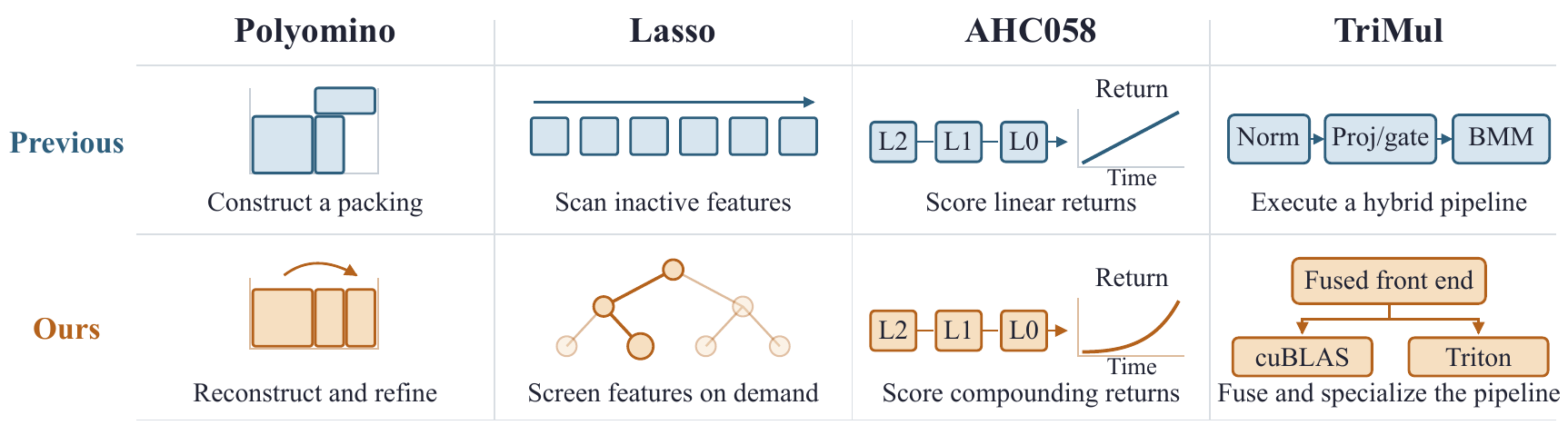}
\caption{Comparison of solution structures. Blue denotes prior approaches; gold denotes ours.}
\label{fig:solution-mechanisms}
\end{figure}

Across all four tasks, Guidance-TTT exhibits a common pattern: it preserves a strong algorithmic backbone while introducing \textbf{targeted structural changes} to how search, computation, or adaptation is allocated (Figure~\ref{fig:solution-mechanisms}). These comparisons are post hoc: Guidance-TTT does not directly evolve the referenced human or prior solutions. Rather, we compare the independently discovered solutions with strong existing approaches to characterize the structural changes associated with their gains.  

For Polyomino, the discovered solver independently recovers a geometry-aware skyline construction similar to the human solver, while adding a simulated-annealing refinement stage over piece priorities and board width. For Lasso, it combines a homotopy-style path method with segment-tree-based screening, avoiding repeated scans over the full inactive set. For AHC058, it introduces phase-dependent scheduling and cascade-aware valuation, allowing investment priorities to adapt as the production state evolves. For TriMul, it arrives at a hybrid Triton--cuBLAS design with additional front-end fusion and shape-dependent contraction dispatch.  

\subsection{Ablations}
\label{sec:ablations}
Having established the overall discovery performance in Table~\ref{tab:main-results}, we next isolate the effects of key design choices in Guidance-TTT. Table~\ref{tab:ablations} reports matched 8B ablations on Polyomino and TriMul, comparing the full method with frozen guidance, direct solution training, and an extra-parent-context variant (G-Top2). The full four-task guidance-scaling analysis is reported in Appendix~\ref{sec:guidance-scaling}.

\begin{table}[!htbp]
\centering
\caption{Ablations on Polyomino and TriMul.
Bold marks the best result among the matched 8B configurations.}
\label{tab:ablations}

\begin{threeparttable}
\small
\setlength{\tabcolsep}{5pt}
\renewcommand{\arraystretch}{1.08}

\begin{tabularx}{\linewidth}{@{}Xrr@{}}
\toprule
Setting
& Polyomino score $\uparrow$
& TriMul latency ($\mu$s) $\downarrow$ \\
\midrule

\multicolumn{3}{@{}l}{\textit{Matched 8B configurations}} \\[1pt]

Guidance-TTT
& \textbf{91.89}
& 1225.58 \\

Frozen guidance
& 84.85
& 1256.62 \\

Direct solution training
& 41.63
& 9664.91 \\

Extra parent context (G-Top2)
& 82.12
& \textbf{1166.82} \\

\bottomrule
\end{tabularx}

\begin{tablenotes}[flushleft]
\scriptsize
\item[] All matched configurations use Qwen3-8B and the same rollout budget.
\emph{Frozen guidance} keeps the guidance model fixed while retaining GLM-5.2 execution.
\emph{Direct solution training} trains Qwen3-8B to generate complete solutions directly with TTT approach.
\emph{G-Top2} adds a second PUCT-selected parent's summary and score to the guidance context.
\end{tablenotes}

\end{threeparttable}
\end{table}

\noindent\textbf{Ablation results.}
The matched 8B controls isolate three design choices in Guidance-TTT:
whether to learn the guidance policy, where to apply test-time adaptation,
and how much parent context to expose to guidance.

First, freezing the guidance model while retaining the same
guidance--execution hierarchy reduces the Polyomino score from 91.89 to
84.85 and increases TriMul latency from \(1225.58\) to
\(1256.62\,\mu\text{s}\). Therefore, the hierarchical decomposition alone is not
sufficient, updating the guidance policy provides additional gains on both
tasks. Second, applying the same 8B adaptation budget directly to a model that
generates complete solutions performs substantially worse, reaching 41.63 on
Polyomino and \(9664.91\,\mu\text{s}\) on TriMul. This comparison suggests
that concentrating test-time learning on strategic guidance is markedly more
effective than asking the adapted model to both choose and implement complete
solution changes. Third, adding a second parent's summary and score (G-Top2) has opposite
effects across the two tasks: it degrades Polyomino to 82.12 but improves
TriMul to \(1166.82\,\mu\text{s}\). This contrast suggests that additional
cross-parent context is useful only when proposed changes transfer cleanly
across nearby solution states. Broader guidance-scale
and execution-model comparisons are reported in
Appendices~\ref{sec:guidance-scaling} and~\ref{app:executor-comparisons}, respectively.

\noindent\textbf{Practical advantage: reducing adaptation cost by freezing the large executor.}
As an additional large-executor comparison on Polyomino, Guidance-TTT with a
frozen GPT-OSS-120B executor reaches 83.19, close to the 83.72 obtained by
TTT-Discover when directly adapting the same 120B model. Guidance-TTT trains
only Qwen3-8B guidance with \(8\times16\) samples for 30 updates, whereas
TTT-Discover trains GPT-OSS-120B with \(8\times64\) samples for 50 updates.
Under the hardware and service configurations used in these runs, the observed
cost is approximately \$120 for Guidance-TTT versus \$283 for the TTT-Discover
trace, a reduction of roughly 58\%. This result illustrates the practical
cost advantage of keeping a large execution model frozen while retaining
similar solution quality. Full executor comparisons are reported in
Appendix~\ref{app:executor-comparisons}.

\subsection{Analysis: Scaling Guidance Improves Performance on Saturated Tasks}
\label{sec:scaling-guidance-capacity}

In this section, we analyze why scaling guidance improves discovery on saturated tasks. On both AHC058 and TriMul, 14B guidance
improves over the 8B runs and reaches state-of-the-art performance. We compare
the resulting solutions and diagnostic probes to identify what the larger
guidance model does differently and how this translates into better results.
Full trajectories and matched comparisons are reported in
Appendix~\ref{sec:guidance-scaling}.

\begin{figure}[H]
\centering
\includegraphics[width=0.82\linewidth]{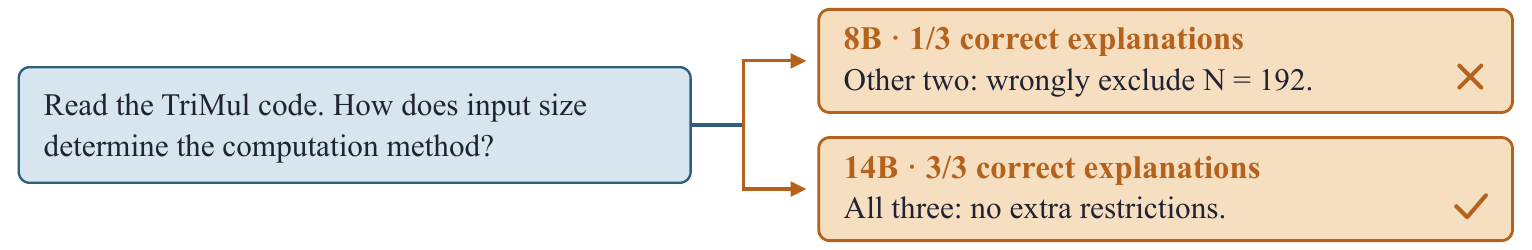}
\caption{TriMul model-capability probe. Counts report explanations without
incorrect extra conditions across three samples per model (higher is better).}
\label{fig:guidance-probe-dialogue}
\end{figure}

The 8B and 14B runs recover similar high-level mechanisms, but 14B applies
them with finer specialization. On TriMul, it adapts the fusion strategy to
channel size rather than using a common pipeline. On AHC058, it combines
state-dependent valuation with beam-based initialization and coordinated
cross-level moves. These solutions indicate that larger guidance better
organizes techniques around the current solution state. To examine their
understanding, we asked two models to explain the SOTA code generated by Guidance-TTT with 14B model and answer related questions. Both recognized the main mechanisms. On the TriMul explanation probe, 8B added incorrect branch restrictions in
two of three responses, while 14B was correct in all three
(Figure~\ref{fig:guidance-probe-dialogue}).
This example illustrates a gap between \textbf{recognizing a mechanism and accurately
describing when it applies.}

\subsection{Analysis: Guidance-TTT Dynamics}
\label{sec:shared-search-dynamic}
\begin{figure}[!htbp]
\centering
\includegraphics[width=\linewidth,trim=0 5bp 0 5bp,clip]{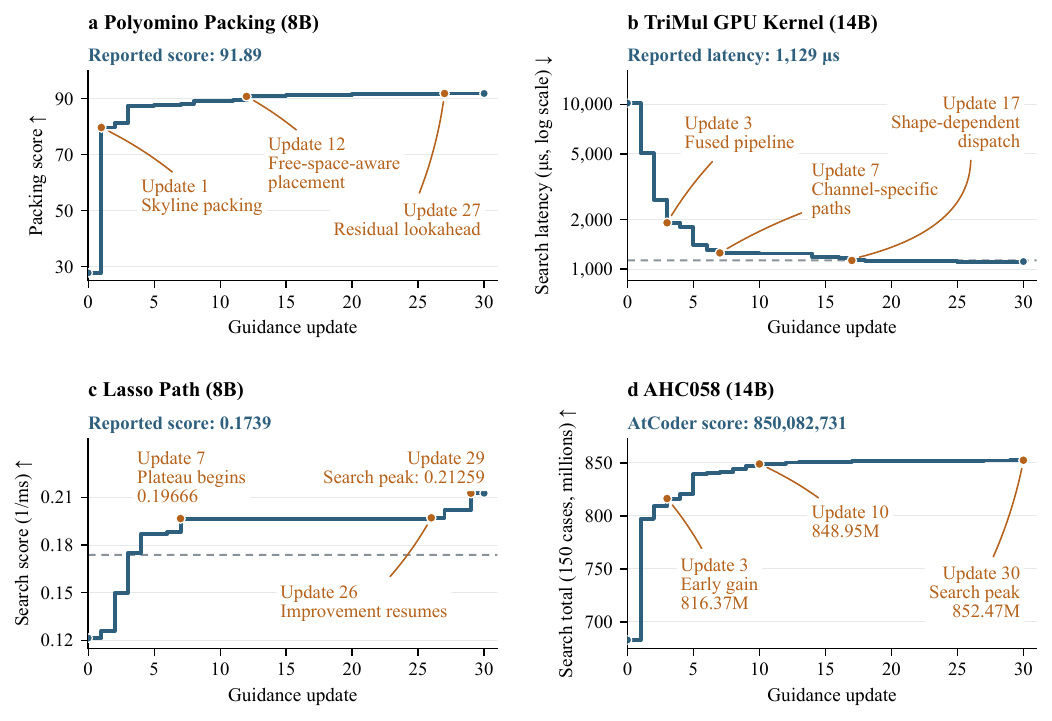}
\caption{Best-so-far search trajectories under GLM-5.2 execution. Search scores and Table~\ref{tab:main-results} results use different evaluation environments.}
\label{fig:best-solution-trajectories}
\end{figure}

Figure~\ref{fig:best-solution-trajectories} shows structural gains followed by
local refinement. Gains also follow plateaus: Lasso improves after updates~7--25,
AHC058 resumes gains late, and TriMul reaches its best latency at update~25.
Update~0 denotes the seed. Further trajectories and lineage analyses appear in
Appendices~\ref{app:guidance-size-stats} and~\ref{app:further-analysis}.

\FloatBarrier
\section{Conclusion}
\label{sec:conclusion}

We introduced \emph{Guidance-TTT}, a test-time RL framework that separates
strategic adaptation from solution implementation. Rather than updating the
model that generates complete solutions, Guidance-TTT learns a compact guidance
policy in \emph{thought space}, while a stronger execution model remains frozen.
Across four distinct discovery domains, this simple separation produces
solutions that outperform the strongest prior-work references while remaining
competitive with public leaderboard results. Controlled comparisons further
show that learning strategic guidance can be substantially more effective than
either freezing the guidance policy or directly adapting a solution-generating
model, while avoiding the cost of training the large execution model.

Guidance-TTT suggests that test-time adaptation can learn to guide a frozen
executor. Future work includes richer guidance representations, improved credit
assignment between guidance and execution, and more diversity-aware search.

\section*{Acknowledgments}
This research is supported by the National Research Foundation (NRF),
Prime Minister's Office, Singapore, under its Campus for Research
Excellence and Technological Enterprise (CREATE) programme.
The Mens, Manus, and Machina (M3S) is an interdisciplinary research
group (IRG) of the Singapore-MIT Alliance for Research and Technology
(SMART) centre.
We sincerely thank Modal Labs for generously providing compute credits
through the Modal for Academics program, and Thinking Machines Lab for
its generous support through a Tinker Research Grant.

\section*{Statement on AI Usage}
We used generative AI tools to assist with code implementation and debugging, experimental analysis and interpretation, literature search and checking, manuscript editing, and figure preparation. AI-assisted code and analyses were reviewed and verified by the authors, and experimental results reported in the paper were obtained from the stated evaluation pipelines rather than generated by AI assistants. The language models that form part of the proposed discovery system and experiments are described separately in the main text and Appendix D. The authors take responsibility for the final text, claims, code, and experimental results.

\Needspace{11\baselineskip}
\section*{Reproducibility Statement.}
We provide the full Guidance-TTT algorithm and training details in Appendix~C,
task-specific experimental configurations and evaluation protocols in Appendix~D,
and selected discovered solutions in Appendix~E.
The implementation and evaluation code are available in the Guidance-TTT
recipe directory of Reef linked in the abstract. The accompanying user
guide\footnote{\url{https://reefinfra.ai/docs/user-guide/recipes/guidance-ttt/}}
provides setup instructions, task configurations, and checkpoint-resume procedures.

\bibliographystyle{iclr2027_conference}
\bibliography{references}

\clearpage
\appendix
\section*{Appendix Overview}
The appendix has five parts. Appendix~\ref{app:further-results} reports additional results, including Lasso
OOD evaluation and TriMul compiler transfer.
Appendix~\ref{app:further-analysis} examines
guidance scaling, parent attachment, and failure modes.
Appendix~\ref{app:method-details} specifies the search and training procedure.
Appendix~\ref{app:settings} gives the experimental settings and evaluation
protocols. Appendix~\ref{app:discovered-solutions} presents selected code.

\FloatBarrier
\Needspace{10\baselineskip}
\section{Further Results}
\label{app:further-results}

We report transfer evaluations of fixed solutions and Polyomino executor
comparisons. The transfer evaluations keep each
selected program fixed and do not use OOD feedback for further search.

\subsection{Lasso OOD Evaluation}
\label{app:lasso-ood}

We evaluate each frozen Lasso solver, without further selection or training, on
two held-out suites. The real-data suite contains four datasets bundled with
SimpleTES: breast cancer, diabetes, DNA, and leukemia. The stress suite contains
11 fixed equicorrelated or dense-signal cases from the SimpleTES evaluation
generator, including feature dimensions up to 50,000. All measurements use
double precision, one CPU thread, three timing repetitions, and the same
\(10^{-6}\) objective-gap tolerance as the main evaluator. We report a suite's
geometric-mean latency only if the solver passes every case in that suite.

\begin{table}[!htbp]
\centering
\caption{Frozen Lasso OOD results. Time is the geometric mean in milliseconds;
a dash indicates that at least one case failed correctness.}
\label{tab:lasso-ood-results}
\small
\setlength{\tabcolsep}{5pt}
\renewcommand{\arraystretch}{1.10}
\begin{tabular}{l c c c c c}
\toprule
Method
& \makecell{Real\\pass}
& \makecell{Real\\time $\downarrow$}
& \makecell{Stress\\pass}
& \makecell{Stress\\time $\downarrow$}
& \makecell{All OOD\\pass} \\
\midrule
SimpleTES public best \citep{ye2026evaluation}
& 4/4 & \textbf{3.370} & 9/11 & -- & 13/15 \\
\rowcolor[gray]{0.93}
Guidance-TTT, Qwen3-8B + GLM-5.2
& 4/4 & 4.003 & \textbf{11/11} & \textbf{8.068} & \textbf{15/15} \\
\rowcolor[gray]{0.93}
Guidance-TTT, Qwen3-8B + GPT-OSS-120B
& 2/4 & -- & 11/11 & 11.828 & 13/15 \\
\rowcolor[gray]{0.93}
Guidance-TTT, Qwen3-14B + GLM-5.2
& 4/4 & 3.479 & 11/11 & 8.535 & 15/15 \\
\bottomrule
\end{tabular}
\end{table}

The two solvers discovered by Guidance-TTT with GLM-5.2 execution pass all 15 OOD cases, whereas
the SimpleTES solver fails two high-dimensional correlated stress cases. The
advantage is robustness rather than uniformly lower latency: SimpleTES is
faster on the four real datasets, while the 8B+GLM solver is fully correct on
the stress suite and has the lowest stress latency among the discovered solvers
that pass every stress case. The GPT-OSS-120B solver fails two real-data cases,
consistent with its weaker stability on the main evaluator.

\noindent\textbf{Extended real-data evaluation.}
We also evaluate the frozen 8B+GLM and 14B+GLM solvers on the full
11-dataset real-data suite in Table~\ref{tab:lasso-real11}.
These measurements are separate from the four-dataset evaluation above;
DNA and leukemia occur in both suites. No OOD feedback is used to train or
select the solvers. All three methods run serially on the same CPU core,
with one BLAS/OpenMP thread and double precision. We use 50 regularization
values with $\lambda_{\min}/\lambda_{\max}=10^{-2}$, one warm-up, and five
timed repetitions, including input copying. We check every invocation
against scikit-learn 1.9.0's \texttt{lasso\_path} with an objective-gap
tolerance of $10^{-6}$ and rotate method order across datasets.

\begin{table}[!htbp]
\centering
\begin{threeparttable}
\caption{Lasso transfer to 11 real OOD datasets. Latency is the mean of five
runs in milliseconds ($\downarrow$); speedup is SimpleTES latency divided by
ours ($\uparrow$). Both solvers were discovered using GLM-5.2 as the executor.
Bold marks the fastest valid solver in each row.}
\label{tab:lasso-real11}
\small
\setlength{\tabcolsep}{5pt}
\renewcommand{\arraystretch}{1.13}
\begin{tabular*}{\linewidth}{@{\extracolsep{\fill}}l r r r r r@{}}
\toprule
& \multicolumn{3}{c}{Latency (ms) $\downarrow$}
& \multicolumn{2}{c}{Speedup $\uparrow$} \\
\cmidrule(lr){2-4}\cmidrule(l){5-6}
Dataset & SimpleTES & Ours, 8B & Ours, 14B & 8B & 14B \\
\midrule
Gisette           & 1,850.35\tnote{$\dagger$} & \textbf{597.56} & 2,111.72 & -- & -- \\
RCV1              & 10,225.08 & \textbf{10,156.79} & 63,674.64 & 1.01$\times$ & 0.16$\times$ \\
DNA               & 10.16 & \textbf{7.50} & 13.68 & 1.35$\times$ & 0.74$\times$ \\
Leukemia          & 10.32 & 9.34 & \textbf{8.05} & 1.10$\times$ & 1.28$\times$ \\
Colon Cancer      & 7.45 & 4.76 & \textbf{4.45} & 1.57$\times$ & 1.68$\times$ \\
Duke Breast Cancer & 11.38 & 9.35 & \textbf{8.44} & 1.22$\times$ & 1.35$\times$ \\
\addlinespace[3pt]
TCGA BRCA         & 1,893.85\tnote{$\dagger$} & \textbf{236.54} & 1,653.82 & -- & -- \\
TCGA Liver RNA    & 186.98 & 143.50 & \textbf{127.51} & 1.30$\times$ & 1.47$\times$ \\
TCGA Lung RNA     & 227.53 & 177.34 & \textbf{174.87} & 1.28$\times$ & 1.30$\times$ \\
TCGA Prostate RNA & 227.37 & 146.43 & \textbf{141.28} & 1.55$\times$ & 1.61$\times$ \\
TCGA Thyroid RNA  & 219.53 & 176.22 & \textbf{147.53} & 1.25$\times$ & 1.49$\times$ \\
\midrule
Correctness pass & 9/11 & 11/11 & 11/11 & & \\
\multicolumn{4}{l}{Geometric-mean speedup (9 mutually valid datasets)} & 1.28$\times$ & 1.05$\times$ \\
\multicolumn{4}{l}{Faster than SimpleTES (9 mutually valid datasets)} & 9/9 & 7/9 \\
\bottomrule
\end{tabular*}
\begin{tablenotes}[flushleft]\footnotesize
\item[$\dagger$] Exceeds the objective-gap threshold under this evaluator;
latency is shown for completeness and excluded from speedup comparisons.
SimpleTES is the public best solver of \citet{ye2026evaluation}.
\end{tablenotes}
\end{threeparttable}
\end{table}

Both discovered solvers pass all 11 datasets. On the nine datasets where
SimpleTES also passes, the 8B solver is faster on all nine, with a
$1.28\times$ geometric-mean speedup; its RCV1 advantage is only $0.7\%$.
The 14B solver is faster on seven of nine, but its RCV1 slowdown reduces
its geometric-mean speedup to $1.05\times$. Thus, the 8B solver transfers
its speed advantage more consistently across this suite, while the larger
guidance model does not produce a uniformly faster solver.
SimpleTES exceeds the objective-gap threshold on Gisette
($1.13\times10^{-6}$) and TCGA BRCA ($8.38\times10^{-6}$).
The reference solver emits convergence warnings on TCGA BRCA and
TCGA Prostate, so correctness counts are specific to this evaluation
protocol rather than a claim of exact optimality.

\subsection{TriMul Compiler Transfer}
\label{app:trimul-compiler-transfer}

We select the 14B-guided kernel on Triton 3.3.1 and evaluate it unchanged
on Triton 3.4.0 and 3.6.0. Table~\ref{tab:trimul-compiler-transfer} retains
the full comparison, including additional automated methods and public
submissions. The published TTT-Discover latency of 1161\,$\mu$s uses its
original GPUMode runtime; its released kernel measures 1262\,$\mu$s in our
Triton 3.3.1 environment. All entries within a column use the same runtime.
The excluded rank-2 submission (ID 460994) and rank-3 submission (ID 781830)
compile custom CUDA/C++ code with \texttt{torch.utils.cpp\_extension.load\_inline}
and link cuBLAS/cuBLASLt. They fall outside our Triton-with-PyTorch
implementation scope, rather than violating the public leaderboard rules.
This restriction applies equally to our solutions and references.
The included rank-1, rank-4, and rank-5 submissions (IDs 781368, 781100,
and 408370) and our kernels use Triton without custom CUDA/C++ extensions.

\begin{table}[!htbp]
\centering
\caption{TriMul geometric-mean latency for fixed kernels across Triton
versions. Our kernel is selected on Triton 3.3.1 and evaluated unchanged on
the OOD versions. GPUMode ranks follow the public leaderboard snapshot cited in this paper
\citep{gpumodetrimul}. Underlining marks the fastest automated discovery
result in each column.}
\label{tab:trimul-compiler-transfer}
\small
\setlength{\tabcolsep}{4.0pt}
\renewcommand{\arraystretch}{1.10}
\begin{tabular}{
    >{\raggedright\arraybackslash}p{0.43\linewidth}
    >{\centering\arraybackslash}p{0.15\linewidth}
    >{\centering\arraybackslash}p{0.15\linewidth}
    >{\centering\arraybackslash}p{0.15\linewidth}}
\toprule
Method / solution
& \makecell{Triton 3.3.1\\{\scriptsize development}\\($\mu$s)}
& \makecell{Triton 3.4.0\\{\scriptsize OOD}\\($\mu$s)}
& \makecell{Triton 3.6.0\\{\scriptsize OOD}\\($\mu$s)} \\
\midrule
\multicolumn{4}{l}{\textit{Automated discovery systems}} \\
\addlinespace[1pt]
\rowcolor[gray]{0.93}
\textbf{Guidance-TTT, Qwen3-14B guidance (ours)}
& \underline{1129}
& \underline{1124}
& \underline{1072} \\
K-Search \citep{cao2026ksearch}
& 1131
& 1169
& 1154 \\
TTT-Discover \citep{yuksekgonul2026learning}
& 1262
& 1229
& 1164 \\
Aster \citep{bicker2026aster}
& 1280
& 1232
& 1212 \\
\addlinespace[2pt]
\multicolumn{4}{l}{\textit{Public GPUMode submissions}} \\
\addlinespace[1pt]
GPUMode rank 1 (ID 781368)
& 1128
& 1112
& 1062 \\
GPUMode rank 4 (ID 781100)
& 1163
& 1166
& 1175 \\
GPUMode rank 5 (ID 408370)
& 1278
& 1293
& 1131 \\
\bottomrule
\end{tabular}
\end{table}

\subsection{Execution-Model Comparisons}
\label{app:executor-comparisons}
The main results hold the executor fixed to GLM-5.2.
Table~\ref{tab:executor-results} compares alternative executors on Polyomino.

\begin{table}[H]
\centering
\caption{Polyomino executor variants with Qwen3-8B guidance.}
\label{tab:executor-results}
\small
\begin{tabular}{lrr}
\toprule
Executor & Updates & Polyomino $\uparrow$ \\
\midrule
GLM-5.2 & 30 & \textbf{91.89} \\
GPT-5.4 & 30 & 85.32 \\
GPT-OSS-120B & 30 & 83.19 \\
\bottomrule
\end{tabular}
\end{table}

All three executor variants use a 30-update budget.
The GPT-OSS-120B result of 83.19 is below TTT-Discover's
83.72. The Polyomino cost comparison in the main text is therefore
a lower-cost, slightly lower-score trade-off.

\FloatBarrier
\begingroup
\let\FloatBarrier\relax
\section{Further Analysis}
\label{app:further-analysis}

We examine guidance scaling, parent attachment, and failure modes across the
four tasks. The analyses use stored candidate histories and fixed-solution
evaluations; individual search trajectories do not establish causal scaling laws.

\subsection{Guidance Model Size}
\label{sec:search-progress}
\label{sec:guidance-scaling}
We compare Qwen3-8B and Qwen3-14B guidance across four tasks with a fixed
GLM-5.2 executor and an \(8\times16\), 30-update budget. We measure
\emph{validity}, the fraction of candidates that pass the verifier, and
\emph{useful yield}, the fraction of all candidates that are valid and improve
their selected parent. Performance uses final-program evaluations for
Polyomino, Lasso, and TriMul, and the best training-archive score for AHC058.

\begin{figure}[!htbp]
\centering
\includegraphics[width=\linewidth]{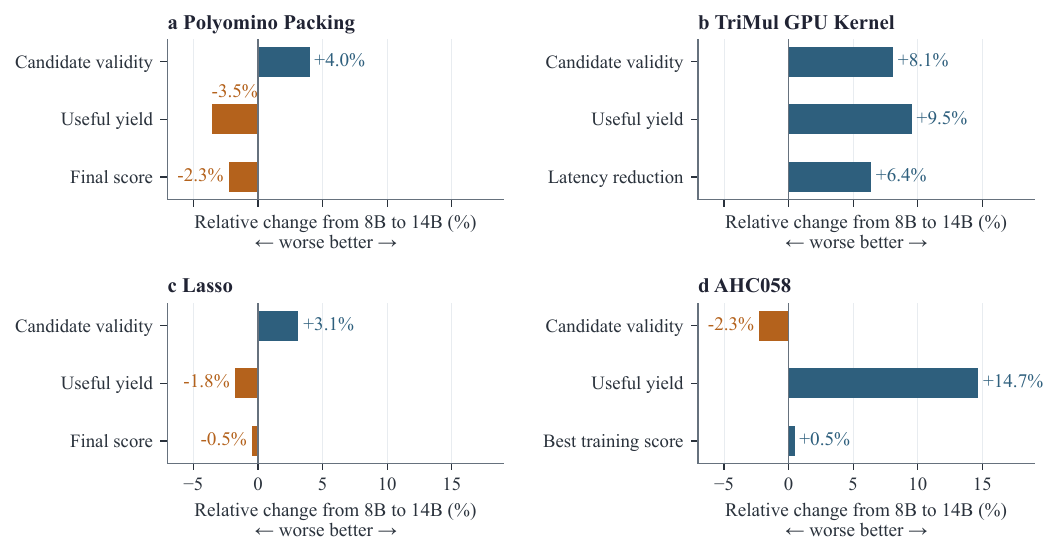}
\caption{Relative change from 8B to 14B guidance with GLM-5.2 execution.
Positive values indicate improvement; TriMul shows relative latency reduction,
and AHC058 compares best training-archive scores. TriMul's 6.4\% reduction
uses the same-batch means of 1297.44\,$\mu$s (8B) and 1214.62\,$\mu$s (14B).
Each size has one search trajectory. Raw validity includes service failures,
which are audited separately below. Absolute statistics are in
Appendix~\ref{app:guidance-size-stats}.}
\label{fig:guidance-scaling}
\end{figure}

\noindent\textbf{Local improvement does not guarantee a better final solution.}
Figure~\ref{fig:guidance-scaling} separates feasibility, parent-relative
improvement, and final performance. TriMul improves on all three measures with
14B guidance. Polyomino has higher validity but lower useful yield and final
score; its 32B run also remains below 8B. Lasso validity rises from 86.48\%
to 89.19\%, while useful yield changes from 7.21\% to 7.08\% and the final
means remain close ($0.1739\pm0.0015$ versus $0.1730\pm0.0023$).
AHC058's raw validity falls from 84.38\% to 82.45\%, while useful yield
rises from 14.56\% to 16.69\% and the best training score increases by 0.46\%.
However, the archives include 95 HTTP~402 failures for 8B and 245 for 14B.
Excluding these requests gives validity of 86.52\% and 88.07\%, respectively.
The raw decline therefore does not establish a feasibility cost of larger
guidance. Removing service failures from the denominator does not restore the
missing candidates or undo any effects on subsequent updates.

\noindent\textbf{A unified interpretation.}
The relevant unit of learning is an \emph{executable, parent-conditioned
change}, not an abstract proposal in isolation. A larger guidance model can
change both the feasibility and usefulness of these actions, but local gains
need not improve the best solution in the archive. Task structure is one
possible explanation: tensor interfaces in TriMul admit component-level
changes, whereas packing changes can depend on coupled ordering, placement,
and repair state. The parent-context ablation is consistent with this account:
adding a second parent's summary improves TriMul useful yield but sharply
reduces Polyomino validity (Appendix~\ref{sec:additional-analysis}). These
observations do not isolate a causal effect of task structure. With one
trajectory per size, search randomness remains an alternative explanation,
and Lasso's small final-mean difference does not establish a reliable ranking.

\subsubsection{Absolute Statistics and Additional Trajectories}
\label{app:guidance-size-stats}
Table~\ref{tab:guidance-size-results} collects the retained 8B and 14B results
behind the main table. The candidate statistics below describe the corresponding
search trajectories.

\begin{table}[htbp]
\centering
\begin{threeparttable}
\caption{Retained results by guidance size across four tasks.}
\label{tab:guidance-size-results}
\small
\setlength{\tabcolsep}{4pt}
\begin{tabularx}{\linewidth}{@{}Xrr@{}}
\toprule
Task & 8B guidance & 14B guidance \\
\midrule
Polyomino $\uparrow$ & \textbf{91.89} & 89.80 \\
Lasso $\uparrow$ & $\mathbf{0.1739}\pm0.0015$ & $0.1730\pm0.0023$ \\
AHC058 $\uparrow$ & 849,063,233 & \textbf{850,082,731} \\
TriMul ($\mu$s) $\downarrow$ & 1225.58 & \textbf{1128.91} \\
\bottomrule
\end{tabularx}
\begin{tablenotes}[flushleft]
\footnotesize
\item[] Both use frozen GLM-5.2 execution. AHC058 reports AtCoder scores.
TriMul's 8B entry is the archive endpoint used in the ablation; its 14B entry
is the retained final-kernel mean (rounded to 1129 in the main table).
Figure~\ref{fig:guidance-scaling} instead uses the two sizes evaluated together
in a separate batch; see Appendix~\ref{app:fixed-evaluation}.
Bold marks the better retained value, not statistical significance.
\end{tablenotes}
\end{threeparttable}
\end{table}

\begin{table}[htbp]
\centering
\caption{Absolute candidate statistics for GLM-5.2 execution. Validity and
useful yield use all 3,840 candidate attempts as denominator, including service
failures. Useful candidates must pass the verifier and strictly improve their
exact selected parent.}
\label{tab:guidance-size-stats}
\small
\begin{tabular}{lrrr}
\toprule
Task & Guidance size & Validity (\%) & Useful yield (\%) \\
\midrule
Polyomino & 8B & 89.48 & 14.09 \\
 & 14B & 93.05 & 13.59 \\
 & 32B & 83.15 & 9.30 \\
TriMul & 8B & 63.80 & 8.70 \\
 & 14B & 68.96 & 9.53 \\
Lasso & 8B & 86.48 & 7.21 \\
 & 14B & 89.19 & 7.08 \\
AHC058 & 8B & 84.38 & 14.56 \\
 & 14B & 82.45 & 16.69 \\
\bottomrule
\end{tabular}
\end{table}

\begin{figure}[htbp]
\centering
\includegraphics[width=\linewidth]{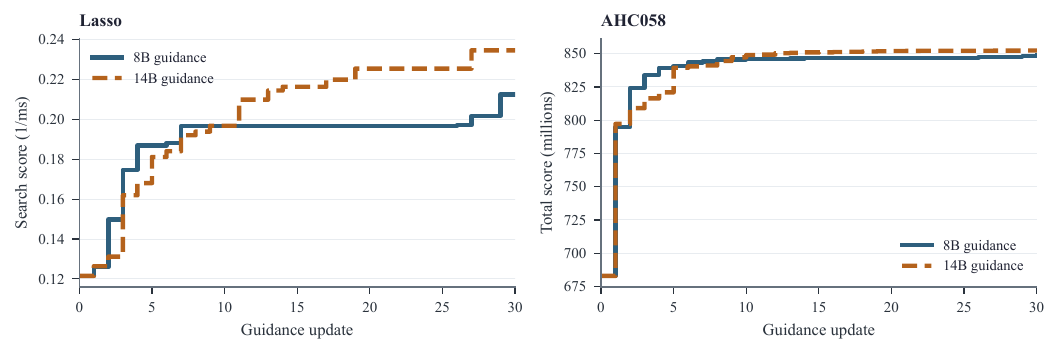}
\caption{Both guidance sizes on Lasso and AHC058, with the same GLM-5.2
seed within each task. These are search-time archive scores. Lasso's 14B
archive peak (0.23467) exceeds the 8B peak (0.21259), but its frozen-solver
mean (0.1730) does not exceed the 8B mean (0.1739).}
\label{fig:extended-size-trajectories}
\end{figure}

The Lasso and AHC058 statistics are reconstructed from the committed
30-update archives, excluding seed entries and uncommitted attempts.
Validity counts agree with the training-step summaries. Useful yield compares
each valid child with its exact parent; it is not the frequency of global
record improvements. These descriptive counts come from one search per
configuration and do not provide between-run uncertainty estimates.

\FloatBarrier
\subsubsection{Diagnostic Probes of Algorithmic Understanding}
\label{app:guidance-knowledge-probes}
To further examine the interpretation in Section~\ref{sec:scaling-guidance-capacity}, we probed Qwen3-8B and Qwen3-14B with a fixed set of prompts, drawing three samples per prompt and model. The probes asked the models to propose improvements, explain excerpts from the 14B solutions, and solve related cases with or without a mechanism hint. In total, we collected 84 responses: 48 with reasoning disabled and 36 with reasoning enabled. For the reasoning-enabled runs, we used temperature~0.6, top-\(p=0.95\), and an 8192-token completion limit.

With reasoning enabled, both models passed all explicit code-condition checks on AHC058 (15/15) and TriMul (18/18). A difference emerged, however, in the quality of the TriMul explanations. Among the three 8B responses, only one correctly described the relevant branch conditions without introducing additional restrictions, whereas all three 14B responses did so correctly. The other two 8B responses incorrectly imposed extra divisibility requirements. In particular, the persistent contraction fallback requires only \(N\bmod128\ne0\), not the additional condition that \(N\) be indivisible by 64 or 32. Because this fallback is inactive on the standard benchmark shapes, these errors cannot directly account for the observed performance gap. They nevertheless suggest that the larger guidance model more reliably captures and articulates the underlying mechanism, consistent with the interpretation that increased guidance capacity improves strategic reasoning rather than merely code-condition compliance.
\subsubsection{Selected Diagnostic Comparisons}
\label{app:guidance-cross-examination}

To examine where smaller guidance can miss an implementation condition, we
select four checks on which 8B makes more errors than 14B
(Table~\ref{tab:guidance-cross-examination}). These are illustrative cases
selected after examining the answers, not a summary of overall diagnostic
accuracy. Both sizes receive
identical code-reading or transfer questions, with source-model identities
and discovery scores hidden. Reference answers are fixed before the calls
and checked against the source or executable examples.

\begin{table}[htbp]
\centering
\caption{Selected checks where 14B answers correctly more often than 8B.
Counts are correct responses out of three to the same prompt; larger is
better. For the AHC058 boundary check, both boundaries must be correct.}
\label{tab:guidance-cross-examination}
\small
\setlength{\tabcolsep}{6pt}
\begin{tabular}{@{}llcc@{}}
\toprule
Task & Check & 8B & 14B \\
\midrule
Polyomino & Score of an exact fit & 1/3 & 2/3 \\
TriMul & Condition for skipping a whole tile & 2/3 & 3/3 \\
AHC058 & Inclusion of exact phase boundaries & 2/3 & 3/3 \\
AHC058 & Reachability of a joint upgrade & 1/3 & 3/3 \\
\bottomrule
\end{tabular}
\end{table}

\noindent\textbf{Polyomino: tracing the scoring function.}
For a \(2\times2\) box that exactly fills a \(2\times2\) gap, the supplied scoring function returns \(0.55\): the fill term is maximized, but the remaining-space usability term is zero. Two of the three 8B responses instead return \(1\), effectively treating complete geometric coverage as equivalent to the full heuristic score. The 14B model computes the correct value in two of three responses. This probe therefore tests whether the model can follow the actual scoring rule rather than rely on an intuitive proxy for solution quality.

\noindent\textbf{TriMul: distinguishing element-level sparsity from tile-level pruning.}
A single zero entry in the mask does not justify skipping an entire tile; tile-level pruning is valid only when all relevant contributions are zero. One 8B response incorrectly promotes an element-level observation into a tile-level execution rule, whereas all three 14B responses reject the shortcut. The distinction is important because exploiting sparsity requires reasoning not only about whether zeros are present, but also about the granularity at which they permit computation to be safely removed.

\noindent\textbf{AHC058: handling boundary cases and coupled decisions.}
The supplied investment rule switches regimes at progress \(0.30\) and \(0.60\). Because the outer branches use strict inequalities, both exact boundary values fall into the state-dependent middle branch. One 8B response misclassifies both boundaries, while all three 14B responses handle them correctly. We also probe a simple transfer example in which the initial reward is \(100\), either upgrade alone yields \(90\), and applying both yields \(160\). Two 8B responses claim that a greedy procedure accepting only individually improving moves can reach the joint upgrade. In fact, neither first move would be accepted, so the higher-reward combination is unreachable under that search rule. All three 14B responses identify this constraint correctly.

Together, these examples probe complementary aspects of mechanism-level reasoning: faithfully evaluating an objective, translating structural observations into valid execution rules, and distinguishing the value of a decision from whether that decision is reachable under the specified search dynamics.

\subsection{Parent Attachment and Transferability}
\label{sec:additional-analysis}

In Guidance-TTT, each guidance is anchored to one selected parent: the guidance model observes
its summaries and score, while the execution model modifies its exact solution.
G-Top2 adds information from a second PUCT-selected solution to the guidance
context, but leaves the executable state attached to the original parent.
Any mechanism borrowed from the alternative must therefore remain compatible
with the selected parent's implementation.

The two tasks respond differently to this intervention. On Polyomino, G-Top2
reduces validity to 33.83\% and useful yield to 7.71\%, indicating that packing
mechanisms depend strongly on the surrounding ordering, state, and repair
logic. On TriMul, validity remains close to the selected-parent setting, useful
yield increases from 8.70\% to 9.17\%, and the run reaches
\(1166.82\,\mu\mathrm{s}\) during search. This contrast is consistent with
the task-structure interpretation in Section~\ref{sec:guidance-scaling}:
TriMul modifications may transfer more readily across nearby solutions,
whereas Polyomino changes may depend more on their parent state. The
single-trajectory comparison does not establish this mechanism causally.

\subsection{Failure Modes}
\label{sec:failure-modes}

The Polyomino and TriMul trajectories reveal two recurring failure modes.
First, verifier reward evaluates
the realized program rather than guidance alone. Therefore, execution failures can
produce misleading learning signals to the guidance model. Second, the interaction between policy
updates and parent selection can concentrate later exploration within a small
number of successful lineages. The Lasso and AHC058 histories contain
counterparts of both structures: related proposals with different implementation
outcomes, and parent-relative improvements that do not advance the global best.

\noindent\textbf{Credit ambiguity.}
Credit ambiguity arises when a zero reward does not reveal whether the failure comes from the guidance or from its realization. Since the verifier scores only the final program, it cannot separate poor guidance from a failed implementation of a sound one. The precondition for this ambiguity---execution failure---is common in our
training logs: 404 of 3,840 Polyomino candidates are invalid (10.5\%), all at compile or parse time, and 1,390 of 3,840 on TriMul (36.2\%), mostly correctness failures. The Polyomino failures are the cleaner evidence here: they occur before any behaviour is observed, whereas a correctness failure does not separate a botched realization from a wrong idea. No group was ever entirely invalid (0 of 240 in both tasks), so each zero-reward failure is compared with valid siblings and has a negative reward-advantage component.

\begin{figure}[H]
\centering
\includegraphics[width=\linewidth]{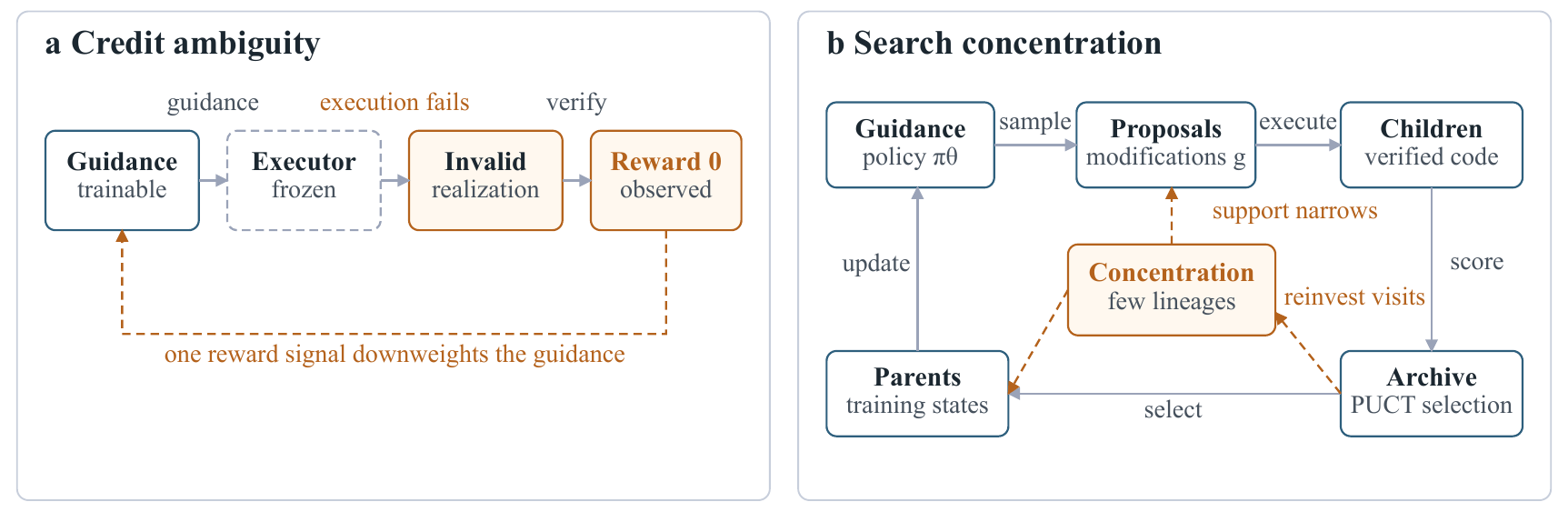}
\caption{Two failure modes in Guidance-TTT. (a) Execution failures can
assign zero reward to useful guidance. (b) The archive--policy loop can
concentrate training and search within a few successful lineages.}
\label{fig:credit-assignment}
\label{fig:search-feedback}
\end{figure}

Polyomino update~21 gives a concrete example. The guidance model proposes
adapting simulated-annealing aspect targets to local packing density, but the
execution model introduces calls before the required placement variables are in scope,
causing compilation to fail. The rollout receives $R=0$ and, against a group mean of \(68.45\) over 16
rollouts (12 valid, 4 invalid, all four compile errors), a reconstructed
reward advantage of approximately \(-1.00\).\footnote{Reward advantages are
recomputed from the 16 stored rewards with the adaptive entropic leave-one-out
estimator in Appendix~\ref{app:rl-update}, giving \(\beta\approx13.14\).
These values exclude the centered base-policy correction; runtime advantage
tensors were not retained.} A second proposal in the same group, from the same parent, asks for the same
mechanism --- density metrics driving an adaptive aspect target --- and the
execution model realizes it without error, producing a valid program scored at
\(91.3173\). This contrast shows why the zero reward is ambiguous: the same underlying mechanism can be realized successfully, so failure to compile does not by itself identify the guidance as poor. The mechanism itself is realizable and was at one point useful: it entered the
run at update~14, improving \(91.0742\) to \(91.1549\), and the resulting
density-weighted aspect update is still present in the update-27 global best
program. On the stronger parents reached by updates~20 and~21, the same
mechanism no longer helped, costing \(0.010\) and \(0.039\) when it was
realized correctly.

The correctly realized variant scored \(0.039\) \emph{below} its parent but
has a reconstructed reward advantage of \(+0.90\), because the estimator
compares siblings rather than improvements over the parent. The group's best
child (\(91.4047\)) has a reward advantage of \(+7.26\): the entropic update
does distinguish among valid solutions. The ambiguity is instead in the
negative feedback on execution failures, which does not identify whether the
proposed strategy or its implementation was responsible.

This ambiguity remains under the entropic objective in
Equation~\ref{eq:guidance-rl}: rewards are observed only after the frozen
execution model realizes the guidance. The objective favors guidance that is
both useful and realizable, but provides no separate label for intrinsic
guidance quality. Comparing siblings does not disentangle these two factors,
since realization difficulty varies within a group.

\noindent\textbf{Trapped in local refinement.}
In our setup, PUCT repeatedly selects strong lineages while policy updates are
trained on the resulting parent distribution, which can gradually concentrate
both guidance and learning around the same solution family. This pattern
appears clearly in Polyomino after the run reaches its global best at
update~27. Across updates~28--30, 341 of 384 candidates remain valid and 17
improve their selected parents, yet none improves the global frontier;
parent-improving children also become rapidly rarer, at 12, 4 and 1 of the
118, 115 and 108 valid children in those updates. All selected parent groups
descend from the same update-20 ancestor, and the median valid child--parent
line-sequence similarity rises from 0.9576 at update~28 to 0.9634 at
update~30, against 0.3195 for randomly paired archive programs under the same
metric. The search therefore keeps refining within one family while the space
of remaining local improvements is itself running out.

\begin{figure}[H]
\centering
\includegraphics[width=\linewidth]{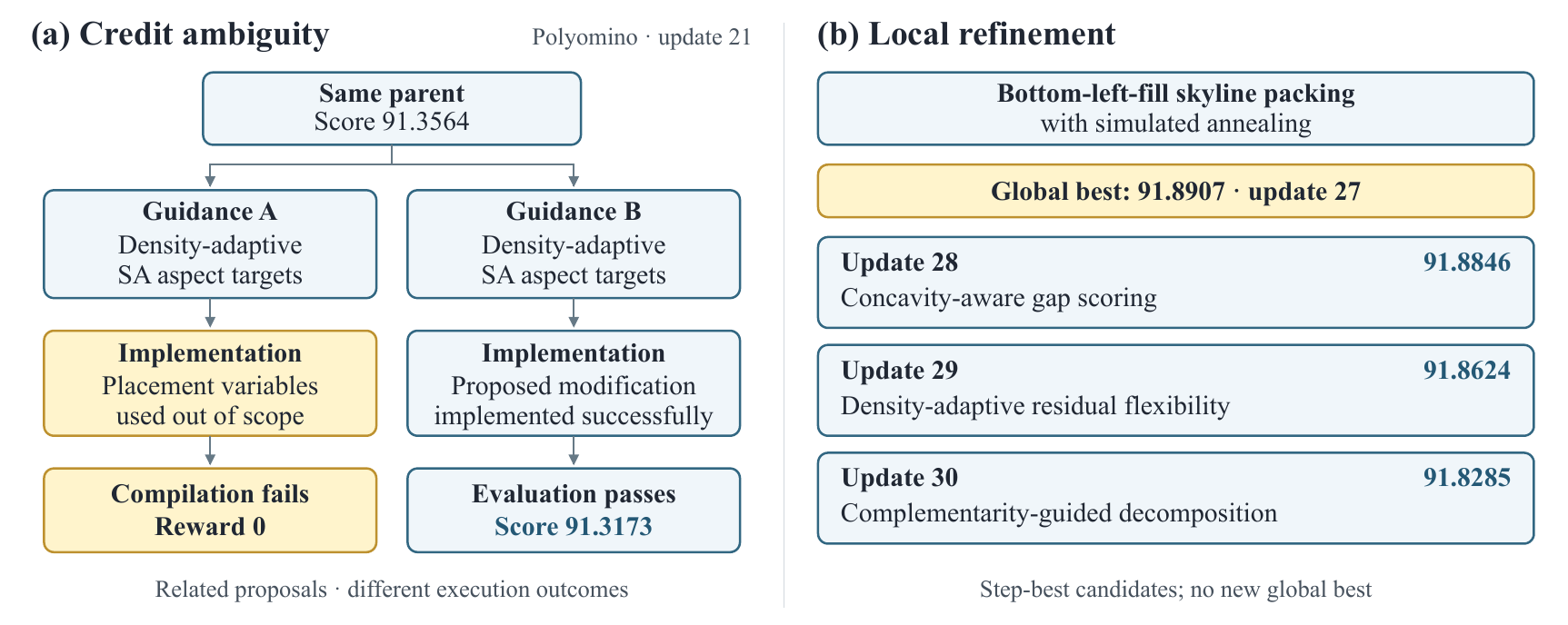}
\caption{Cases corresponding to the two failure modes in
Figure~\ref{fig:credit-assignment}. (a) Two proposals from the same Polyomino
parent receive different learning signals because one execution fails.
(b) The step-best candidates at updates 28--30 add concavity-aware gap
signatures, density-adaptive residual-flexibility weights, and
complementarity-based decomposition feedback, respectively. They retain the
update-20 solver's Bottom-left-fill skyline packing and simulated-annealing
search and do not exceed the update-27 frontier.}
\label{fig:failure-mode-cases}
\end{figure}

TriMul exhibits both patterns as well. At update~2, two proposals from the
same parent recommend a persistent BMM with \(128\times128\times32\) tiles for
large \(N\). One realization fails the public correctness tests and receives
zero reward. A closely related
realization passes the tests and reduces runtime from \(7581.12\) to
\(7470.42\,\mu\mathrm{s}\). The failed realization does not establish that its
guidance was unhelpful because the verifier observes only the resulting
program.

After TriMul reaches its global best of \(1110.85\,\mu\mathrm{s}\) at
update~25, updates~26--30 produce 530 valid candidates, including 24 that
improve their selected parent, without improving the global frontier. All 40
selected parent groups in these updates descend from the same update-14
ancestor. The step-best candidates at updates~28--30 retain the same fused
TriMul pipeline and hybrid BMM dispatch, then adjust tiling and dispatch rules
for narrower sequence-length regimes.

\begin{figure}[H]
\centering
\includegraphics[width=\linewidth]{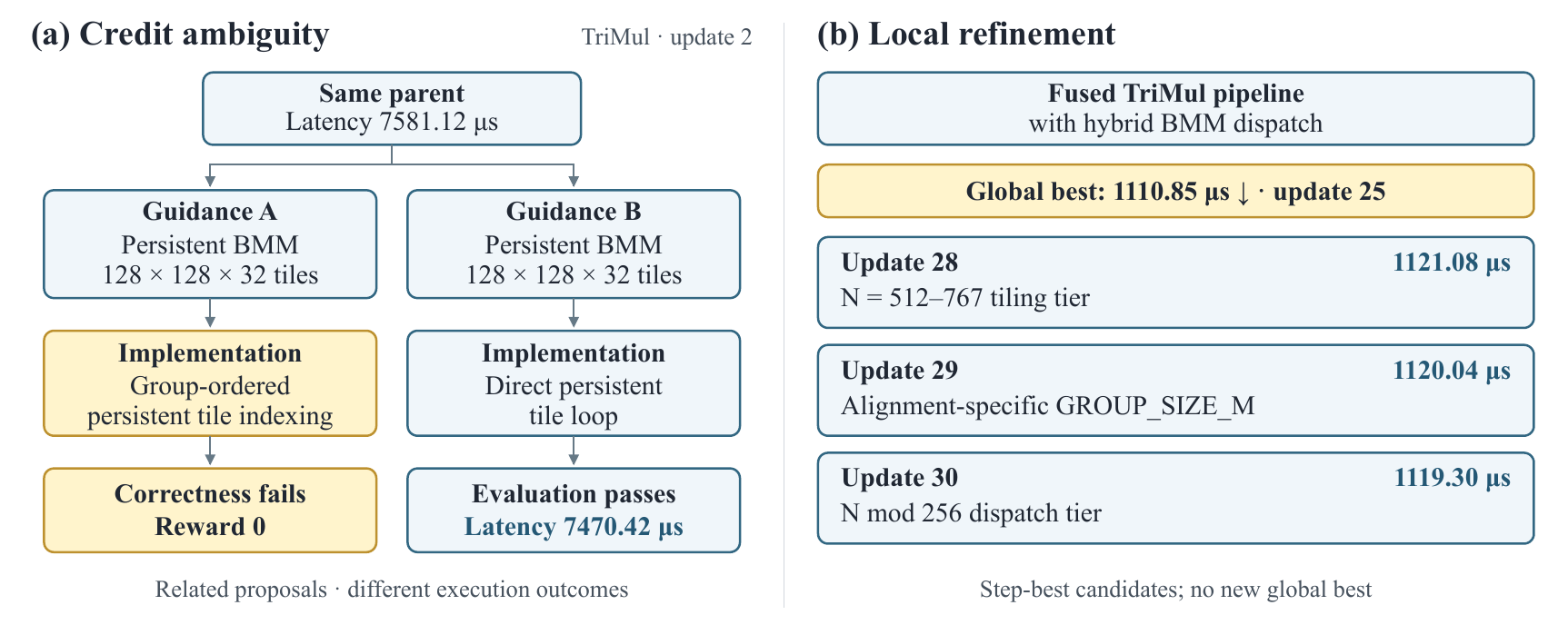}
\caption{TriMul cases corresponding to the two failure modes in
Figure~\ref{fig:credit-assignment}. (a) Two proposals from the same parent use
the same large-\(N\) persistent-BMM schedule, but only one realization passes
correctness tests. (b) After the global frontier stops improving at update~25,
the step-best candidates at updates~28--30 continue to refine BMM tiling and
dispatch within the same kernel family.}
\label{fig:trimul-failure-mode-cases}
\end{figure}

Together, the two task trajectories show that TTT-style updates can reinforce
a strong lineage even after global progress has stopped. PUCT keeps selecting
its descendants, and the guidance policy continues to learn from them,
producing further local improvements without improving the global frontier.
Maintaining diversity across solution families may therefore help the search
escape local optima.

\noindent\textbf{Lasso: the same optimization direction, different realizations.}
At update~19 of the 14B run, two proposals from the same parent request
cache-line alignment of arrays used by the LARS hot path. The parent scores
0.20246. One execution adds an \texttt{ASSUME\_ALIGNED} macro around
\texttt{\_\_builtin\_assume\_aligned}, then assigns its \texttt{void*} result
to typed pointers without a cast. The stored compiler diagnostic identifies
this conversion error; the candidate receives zero reward. Another execution
uses a 64-byte aligned allocator and aligned storage for the active-set
inverse, passes evaluation, and scores 0.22552. Both candidates belong to the
same 16-rollout group, of which 15 are valid
(Figure~\ref{fig:lasso-failure-mode-cases}a).

The paired proposals are not identical, and the observed timing gain does not
isolate alignment's contribution. Nevertheless, the source and diagnostic
establish a concrete implementation failure, while the accepted counterpart
shows that the shared optimization direction can be realized. A zero reward
alone cannot distinguish this type error from an unhelpful alignment proposal.

\noindent\textbf{Lasso: continued refinements within one solver family.}
After the 14B run reaches 0.22552 at update~19, all 56 selected parent groups
in updates~20--26 descend from one update-9 ancestor. The interval produces
776 valid candidates and 37 parent improvements, but no global improvement.
The step-best programs at updates~24--26 retain the LARS plus
coordinate-descent structure and add, respectively, greedy coordinate
selection, lazy Gram-column computation, and Gap Safe screening
(Figure~\ref{fig:lasso-failure-mode-cases}b). Their scores are 0.22164,
0.21691, and 0.22101, below the retained frontier.

This is a temporary concentration of search within one family, not evidence
that further learning is useless: the frontier improves again at update~27.
Nor does a shared ancestor imply identical programs. The ancestry and inspected
changes support local refinement during this interval, while timing variation
limits conclusions about small score differences.

\begin{figure}[htbp]
\centering
\includegraphics[width=\linewidth]{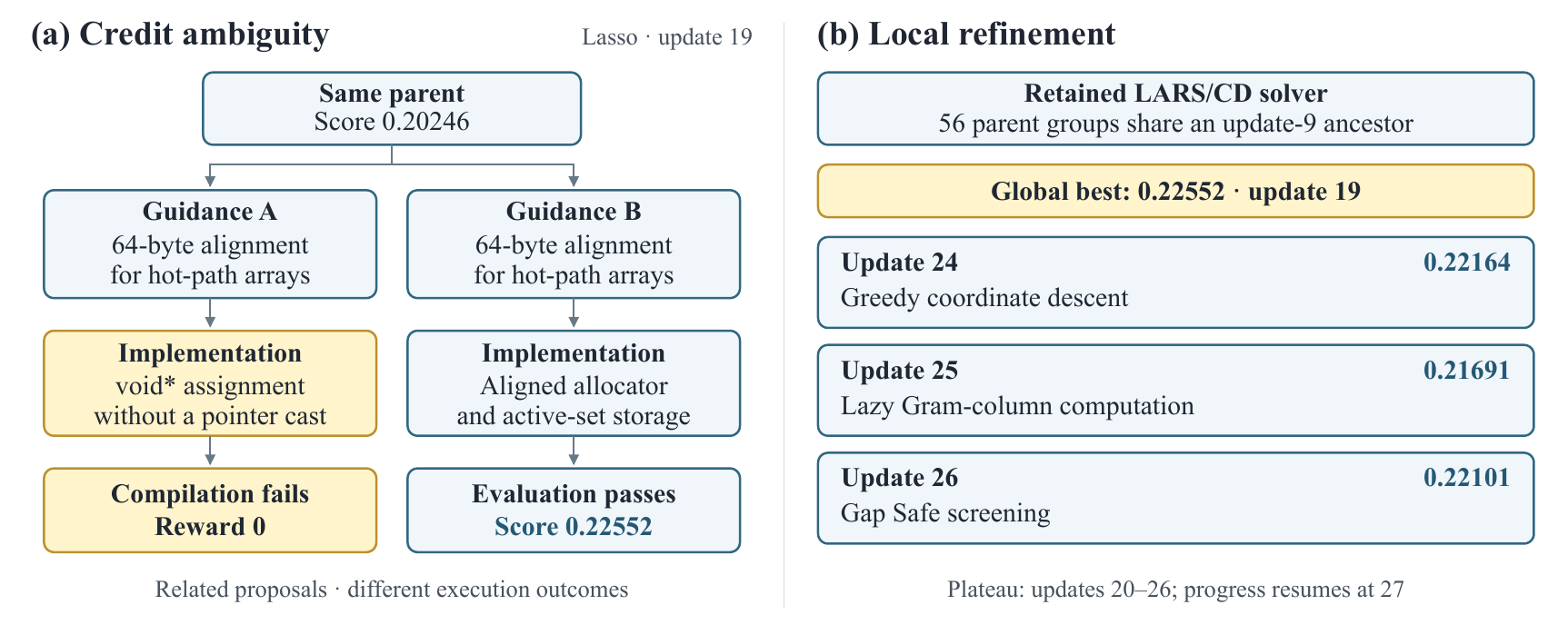}
\caption{Lasso examples of the same two failure structures studied in
Polyomino and TriMul. (a) Related alignment proposals from one parent have
different outcomes because one implementation contains a pointer-conversion
error. (b) During a seven-update plateau, all selected parents share an
update-9 ancestor; the displayed step-best programs refine the retained
LARS/CD solver without exceeding the update-19 frontier. Values are
search-time scores, not fixed-program re-evaluation means.}
\label{fig:lasso-failure-mode-cases}
\end{figure}

\noindent\textbf{AHC058: a declaration error masks a realizable planning change.}
At update~5 of the 14B run, two proposals from the same parent ask for
state-aware phase thresholds and hierarchy-aware lookahead. The parent has
a mean case score of 5,466,338.48. One execution introduces
\texttt{EMAState::update(const State\&,\ldots)} before declaring
\texttt{State}, and accesses its fields in that early definition.
The archive records a compilation failure and zero reward. A syntax-only
compilation of the saved source reproduces the undeclared-type error.

The paired execution derives phase thresholds from a preliminary simulation
and implements level-dependent lookahead. It compiles, passes all 150
instances, and achieves a mean case score of 5,467,815.79
(Figure~\ref{fig:ahc058-failure-mode-cases}a). Its training reward is 1.82261,
using the evaluator's divisor of three million. The group contains 15 valid
programs and one compilation failure. The implementations differ, so this is
not an identical-guidance replay. It nevertheless demonstrates why failure to
compile cannot by itself discredit the shared planning direction.

\noindent\textbf{AHC058: a temporary plateau despite useful local changes.}
At updates~23--26, all 32 selected parent groups descend from one update-18
ancestor. There are 311 valid candidates and 25 parent improvements, but none
exceeds the update-22 frontier of 5,681,246.02 mean points. The step-best
programs at updates~24--26 retain the greedy/beam, simulated-annealing, and
hill-climbing pipeline. They modify hierarchical impact valuation,
phase-dependent level weights, and cascade-aware valuation and moves,
respectively (Figure~\ref{fig:ahc058-failure-mode-cases}b).

The interval also contains 167 HTTP~402 failures, reducing the number of
realized candidates. The plateau therefore cannot be attributed entirely to
the archive--policy interaction. Still, the accepted programs provide direct
evidence of continued local refinement without a frontier improvement.
Progress resumes at update~27 and continues through update~30. As in Lasso,
the observed failure structure is a temporary mismatch between local
improvement and global progress, not demonstrated terminal collapse.

\begin{figure}[htbp]
\centering
\includegraphics[width=\linewidth]{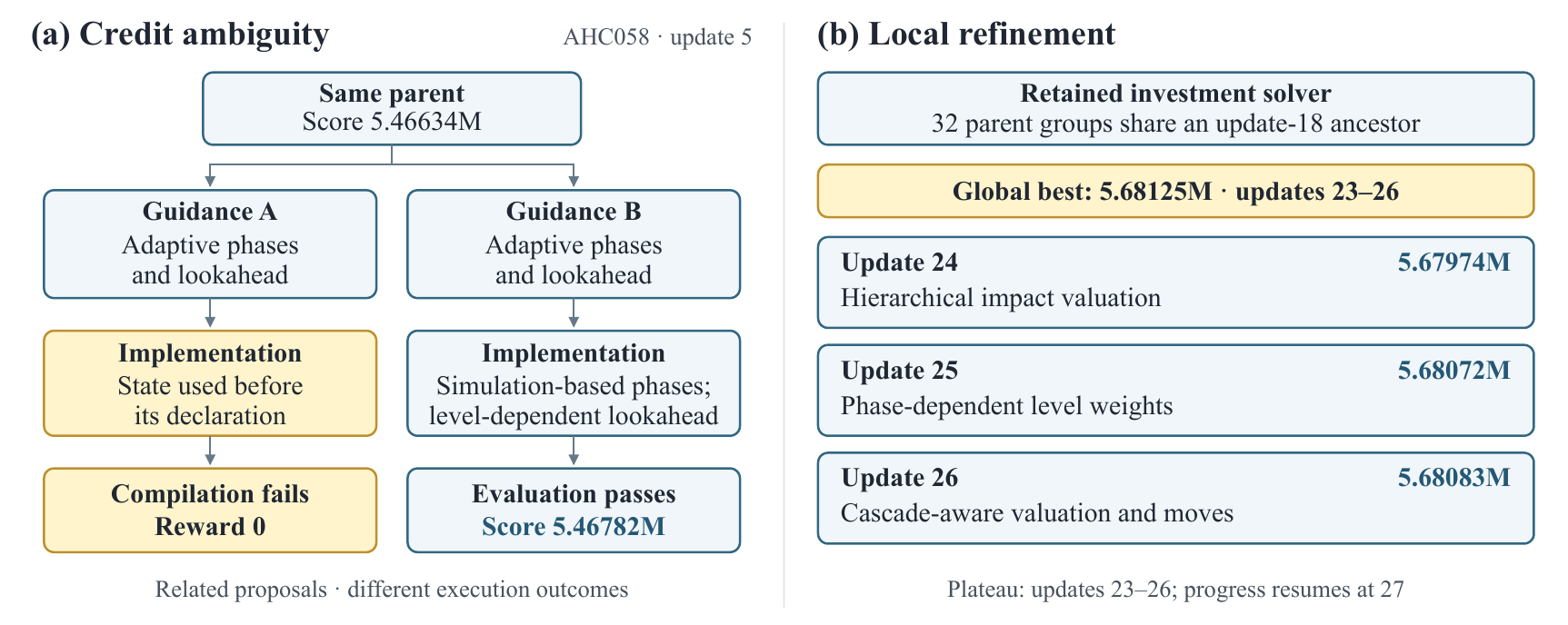}
\caption{AHC058 counterparts of the Polyomino/TriMul failure cases.
(a) Related proposals request adaptive investment phases and lookahead;
one execution fails on a declaration-order error while another passes the
judge. (b) During updates 23--26, selected parents share an update-18 ancestor
and useful local changes do not improve the frontier. Displayed scores are
means over 150 public instances, in millions, rather than the totals used
in the main results. Both panels use the 14B-guidance run.}
\label{fig:ahc058-failure-mode-cases}
\end{figure}

Across these cases, verifier outcomes conflate proposal quality and execution
quality, while parent improvement need not change the global best.
The examples establish these observable structures. They do not separate
the causal effects of policy updates, parent selection, evaluation variation,
and service interruptions. The accompanying audit retains candidate IDs,
parent IDs, group rewards, source hashes, and the ancestry checks.


\endgroup
\clearpage
\section{Method Details}
\label{app:method-details}

\subsection{Budget and Parent Groups}
We use \(U\) for the number of guidance updates and \(N\) for the number of
search candidates. Each update contains \(B\) parent groups and \(K\)
rollouts per group, giving \(N=UBK\). The default \(U=30\), \(B=8\), and
\(K=16\) therefore give 3,840 candidate attempts. Seed generation and its
validation precede this budget; subsequent fixed-solution evaluations are
also separate. A failed candidate still consumes one search attempt.

Parent selection returns an ordered list \((p_1,\ldots,p_B)\), not a set of
necessarily distinct solutions. Parents are selected from the archive visible
before the current update; newly generated children become available at the
next update. The selector prefers nodes outside the ancestor--descendant
lineages already selected in the same batch. If all eligible nodes are
blocked, it falls back to the highest-ranked node. Thus a single seed can
supply all \(B\) groups at initialization, and repeated parents remain
possible later. Each group has its own identifier even when parents repeat.

For group \(b\), the guidance policy samples \(K\) actions conditioned on
the same parent state. Each action is executed once to produce one candidate,
which is then evaluated. These \(K\) candidates form the sibling group for
the advantage estimator; we do not execute one shared guidance \(K\) times.
Independent samples need not produce distinct guidance text or code.

\subsection{Training Records and Search Archive}
Every completed attempt retains its parent link, guidance, generated code,
summaries, verifier status, and reward in the rollout record. The search
archive is a separate, filtered collection used for parent selection. Admission
requires a valid verifier status and an available score; duplicate programs
are filtered. Compilation errors, crashes, and timeouts do not create eligible
parents. For AHC058, all 150 instances must be accepted, even though an invalid
candidate can receive partial training reward. Training therefore uses the
recorded sibling rewards, not only the children admitted to the archive.

TriMul's two-child rule retains the two highest-valued eligible children per
parent for future selection. It does not reduce a group from 16 training
rollouts to two. After a batch, the implementation applies child filtering,
deduplication, and the configured archive-capacity limit. Ancestors required
to preserve lineage links may remain as structural nodes without being
selectable. Pruning the search archive does not delete the full rollout
records or their evaluation feedback.

\subsection{Task Scores and Rewards}
All training rewards are oriented so that larger is better. For valid
candidates, the mappings are:
\begin{itemize}
    \item \textbf{Polyomino:} \(R=s\), where \(s\) is the aggregate packing
    score on the \(0\)--\(100\) scale.
    \item \textbf{Lasso:} \(R=1/t_{\mathrm{geo}}\), where
    \(t_{\mathrm{geo}}\) is geometric-mean solve time in milliseconds across the
    search cases, subject to the objective-accuracy checks.
    \item \textbf{TriMul:} \(R=1500/\ell_{\mathrm{geo}}\), where
    \(\ell_{\mathrm{geo}}\) is geometric-mean latency in microseconds.
    Tables report latency rather than this reciprocal reward.
    \item \textbf{AHC058:} \(R=\bar s/(3\times10^6)\), where \(\bar s\)
    is the evaluator's arithmetic-mean raw score over 150 instances.
    Training-trajectory totals are \(150\bar s\); official AtCoder scores
    are reported separately.
\end{itemize}
Invalid Polyomino, Lasso, and TriMul programs receive zero reward. AHC058
can retain a partial reward from its evaluator, but only fully accepted
candidates contribute archive bests. These task mappings are distinct from
the within-group transformation used to obtain learning advantages.

The Discover-compatible PUCT implementation uses the raw score oriented for
maximization: the packing score, Lasso inverse time, and AHC mean score, or
negative latency for TriMul. This search value is denoted \(v(i)\) in the
main text. In particular, TriMul's PUCT value and its reciprocal RL reward
have the same ordering but are not numerically interchangeable.

\subsection{Adaptive Entropic Advantage}
\label{app:rl-update}
For each selected parent, let \(r_1,\ldots,r_K\) be the realized solutions'
verifier rewards. Following \citet{yuksekgonul2026learning}, we select an
inverse temperature \(\beta\) using
\begin{equation}
q_i(\beta)=\frac{\exp(\beta r_i)}{\sum_{j=1}^{K}\exp(\beta r_j)},
\qquad
D_{\mathrm{KL}}\!\left(q(\beta)\,\|\,\mathrm{Uniform}(K)\right)=\log 2.
\end{equation}
The implementation uses a bounded numerical search, retaining its upper
bound if the target cannot be reached. Constant-reward groups have zero
reward advantage and are masked out of optimization. For nonconstant groups,
the leave-one-out estimator is
\begin{equation}
u_i=\exp\!\left(\beta(r_i-r_{\max})\right),
\qquad
A_i^{\mathrm{reward}}
=\frac{u_i}{\frac{1}{K-1}\sum_{j\ne i}u_j+\epsilon}-1,
\label{eq:entropic-advantage}
\end{equation}
where \(r_{\max}=\max_j r_j\) and \(\epsilon=10^{-12}\).
The sampled rewards and \(\beta\) are held fixed during the policy update.
This estimator compares a candidate with its siblings, not with its parent,
and is not a z-score normalization.

\subsection{Actor Loss and Update Schedule}
Let \(\pi_{\mathrm{rollout}}\) be the guidance policy used to sample the
current batch and \(\pi_{\mathrm{base}}\) the fixed pre-adaptation guidance
policy. The base policy is not the execution model. For response token \(t\)
of guidance \(g_i\), define
\begin{equation}
d_{it}=\log\pi_{\mathrm{rollout}}(g_{it}\mid s_p,g_{i,<t})
-\log\pi_{\mathrm{base}}(g_{it}\mid s_p,g_{i,<t}),
\qquad
A_{it}=A_i^{\mathrm{reward}}+0.1(\bar d-d_{it}),
\end{equation}
where \(\bar d\) is the mean over active response tokens in the batch.
The advantages are held fixed during optimization. Reconstructed reward
advantages in the failure analysis exclude this token-level correction.

The token-mean actor loss is
\begin{equation}
\mathcal{L}(\theta)=
-\frac{\sum_{i,t}m_{it}\,w_{it}\,A_{it}
\log\pi_\theta(g_{it}\mid s_p,g_{i,<t})}{\sum_{i,t}m_{it}},
\end{equation}
where \(m_{it}\) masks padding, prompt tokens, and excluded groups.
The REINFORCE importance weight \(w_{it}\) is detached from differentiation;
when rollout log probabilities are used, it is
\(\operatorname{stopgrad}(\pi_\theta/\pi_{\mathrm{rollout}})\) at that token,
and it is one when no correction is needed. This is not a clipped PPO loss.
No separate actor-KL penalty is added to the centered base-policy term.
An entirely masked batch contributes zero loss.

Each complete \(BK\)-rollout batch supplies one optimizer update, with one
pass through the batch and gradient accumulation over hardware-sized
microbatches. We do not replay the batch for multiple optimization epochs.
Only guidance LoRA parameters are updated; the executor remains frozen.
The Discover-compatible optimizer uses Adam with learning rate
\(4\times10^{-5}\), moments \((0.9,0.95)\), and zero weight decay.
The updated guidance policy is used for the next rollout batch.

\clearpage
\section{Experimental Setup}
\label{app:settings}

\subsection{Primary Configurations}

\begin{table}[H]
\centering
\caption{Primary configurations for Polyomino Packing and TriMul. The Polyomino execution API
had no client-side completion cap. Every Guidance-TTT result on TriMul uses the
\(8\times16\) rollout shape.}
\label{tab:primary-config}
\footnotesize
\setlength{\tabcolsep}{3pt}
\renewcommand{\arraystretch}{1.15}
\begin{tabular}{
    >{\raggedright\arraybackslash}p{0.16\linewidth}
    >{\raggedright\arraybackslash}p{0.35\linewidth}
    >{\raggedright\arraybackslash}p{0.35\linewidth}}
\toprule
Component & Polyomino & TriMul \\
\midrule
Guidance model
& \texttt{Qwen/Qwen3-8B}, LoRA rank 32
& \texttt{Qwen/Qwen3-14B}, LoRA rank 32; 8B for ablations \\
Execution model
& \texttt{z-ai/glm-5.2}, OpenRouter, high reasoning
& \texttt{z-ai/glm-5.2}, OpenRouter, high reasoning \\
Hardware
& 1 B200 for guidance RL; hosted execution
& 1 B200 for guidance RL; isolated H100 verifiers \\
Rollout shape
& \multicolumn{2}{>{\raggedright\arraybackslash}p{0.72\linewidth}}{8 parents/update \(\times\) 16 rollouts/parent} \\
Sampling
& Guidance temperature 1.0, top-\(p=1.0\); provider high-reasoning execution
& Guidance temperature 1.0, top-\(p=1.0\); high-reasoning execution \\
Actor limits
& 4,096 prompt; 8,192 response tokens
& 6,144 prompt; 10,240 response tokens \\
Optimization
& \multicolumn{2}{>{\raggedright\arraybackslash}p{0.72\linewidth}}{Adam, learning rate \(4\times10^{-5}\); adaptive entropic leave-one-out advantage} \\
Base-policy control
& \multicolumn{2}{>{\raggedright\arraybackslash}p{0.72\linewidth}}{Centered log-probability term, coefficient 0.1; no additional actor-KL loss} \\
Communication
& \multicolumn{2}{>{\raggedright\arraybackslash}p{0.72\linewidth}}{Solution summary, parent-relative change summary, and score to guidance; exact solution and guidance to execution} \\
Archive selection
& Maximum-child PUCT, \(c=1.0\)
& Maximum-child PUCT, \(c=1.0\), two retained children/parent \\
Verifier
& FrontierCS problem 0; 70 cases; C++17; 2s/case
& 18 correctness + seven timing cases; 520s timeout \\
Software
& FrontierCS C++17 environment
& CUDA 12.8, PyTorch 2.7.1, Triton 3.3.1 \\
Concurrency
& 16 execution/verification slots
& 16 execution requests; 16 isolated H100 verifiers \\
Reported horizon
& 30 updates; best first reached at update 27
& 30 updates; search-time best first reached at update 25 \\
\bottomrule
\end{tabular}
\end{table}

\begin{table}[H]
\centering
\caption{Configurations for the two additional tasks. Runs use eight parents
per update and 16 rollouts per parent, with a 30-update budget. The AHC058
120B-executor run stopped after 28 updates.}
\label{tab:extended-config}
\footnotesize
\setlength{\tabcolsep}{4pt}
\renewcommand{\arraystretch}{1.12}
\begin{tabular}{>{\raggedright\arraybackslash}p{0.13\linewidth}
                >{\raggedright\arraybackslash}p{0.31\linewidth}
                >{\raggedright\arraybackslash}p{0.25\linewidth}
                >{\raggedright\arraybackslash}p{0.21\linewidth}}
\toprule
Task & Evaluator & Model configurations & Initialization \\
\midrule
Lasso & 17-case search evaluator; objective tolerance \(10^{-6}\)
& Qwen3-8B + GLM-5.2; Qwen3-8B + GPT-OSS-120B; Qwen3-14B + GLM-5.2
& Executor-generated one-shot seed \\
AHC058 & Total raw score over 150 public instances; 2s per instance
& Same three configurations
& Executor-generated one-shot seed \\
\bottomrule
\end{tabular}
\end{table}

Search, reward mappings, and the actor update are specified in
Appendix~\ref{app:method-details}.

\subsection{AHC Evaluation Protocol}

The AHC evaluator executes every candidate on the 150 public instances
released with TTT-Discover and reports their arithmetic-mean raw score. We
convert this mean to the total score over 150 instances for the search
trajectories.\footnote{These training-evaluation totals are distinct from the
official AtCoder submission scores reported in Tables~\ref{tab:main-results}
and~\ref{tab:guidance-size-results}.} A
candidate is eligible for the solution archive only when all 150 instances are
accepted. During training, rejected candidates can still receive a partial
reward for policy optimization, but their scores are never reported as archive
bests. AHC058 evaluates four candidates concurrently with eight case workers per
candidate. The controlled OpenEvolve run starts from the same GLM-5.2 one-shot
program as the corresponding Guidance-TTT condition and uses 500
iterations.

\subsection{Fixed-Solution Evaluation and Transfer}
\label{app:fixed-evaluation}

Lasso's main result freezes the selected program and repeats the 17-case
evaluator five times with newly generated inputs, using one CPU thread on an
AMD EPYC 9745 host. The reported mean includes
zero scores for invalid invocations. The OOD evaluations use fixed programs
without further training or selection. The original real/stress suites and
the extended 11-dataset suite have different timing protocols, which are
specified alongside their results in Appendix~\ref{app:lasso-ood}; they must
not be pooled as independent datasets because DNA and leukemia overlap.

TriMul uses three evaluations per fixed kernel on an isolated H100 80GB GPU,
with CUDA 12.8, PyTorch 2.7.1, and Triton 3.3.1. Each invocation passes 18
correctness tests and reports the geometric mean of seven case latencies;
tables average these three values. The headline ours/K-Search comparison
uses the retained paired batch, with alternating order in one container.
Other public references retain their three-evaluation records under the same
runtime. We also completed a separate batch with seven programs, one container,
and rotating order; the size analysis uses the two guidance sizes measured
together in this separate batch.
The compiler-transfer experiment changes Triton from 3.3.1 to 3.4.0 or 3.6.0
without modifying the selected program. These are software-environment shifts,
not new input distributions. Full results appear in
Appendix~\ref{app:trimul-compiler-transfer}.

Historical controls are kept separate from these final-program measurements.
The retained TriMul OpenEvolve endpoint in Table~\ref{tab:main-results}
and the 8B, frozen-guidance, and G-Top2 endpoints in Table~\ref{tab:ablations}
are archive measurements, not
three-evaluation means. In particular, the 8B/frozen comparison uses the two
archived endpoints. The direct-8B and CORAL rows retain their previously
reported values without a new repeat estimate. The Lasso TTT-Discover value is our earlier measurement;
its repeat vector is not retained in this report. The Lasso
OpenEvolve endpoint likewise retains its original measurement. These historical entries do not
support matched-repeat uncertainty estimates.

AHC058 main results use official-platform scores. TTT-Discover's original
Table~6 reports 848,414,228 points \citep{yuksekgonul2026learning}; SimpleTES's
Table~9 reports 849,325,750, the maximum of ten independent submissions
\citep{ye2026evaluation}. Our entry, 850,082,731, is the recorded score of our
official submission.
These entries are reported submission scores, not
means from a matched repeated-submission protocol.

Standard deviations for Lasso and TriMul describe fixed-program variability,
not variation across training seeds.

\clearpage
\section{Discovered Solution Code}
\label{app:discovered-solutions}

The listings retain the parts of each discovered solution that implement the
design choices discussed below. They omit input handling,
bookkeeping, repeated launch code, and helper functions that do not affect the
mechanisms described here. The accompanying code repository contains the full
implementations; the excerpts are lightly condensed for typesetting.

\subsection{Relationship to Prior Solvers}

\noindent\textbf{Polyomino packing.}
At a high level, the discovered solver is not a complete departure from prior
packing methods, but a \textit{substantial extension of the constructive approach} used
by the released FrontierCS human solver \citep{mang2025frontiercs}. Both
solvers use skyline geometry and search over orientations and widths.
The human solver selects the next piece from a bounded candidate pool during
construction; our solver additionally revises piece priorities and width
through simulated annealing over reconstructed packings.
The main difference from the strongest alternative configurations is a \textit{second-stage refinement procedure}. Guidance-TTT augments the constructive packer with simulated annealing \citep{kirkpatrick1983annealing} and timed restarts, allowing improvement beyond the initial packing. While the strongest frozen and direct-search baselines recover similar constructive backbones, they do not retain this refinement stage. The resulting solver therefore follows a clear two-stage design: construction followed by refinement.

\noindent\textbf{TriMul GPU kernel.}
The discovered kernel \textit{shares the hybrid Triton--cuBLAS structure} used by K-Search \citep{cao2026ksearch}; this is a comparison of final designs, not inherited source code. K-Search combines Triton kernels with cuBLAS for the main computation, whereas our kernel further specializes the pipeline across channel regimes and sequence shapes. In particular, it uses separate front ends for \(C=128\) and \(C=384\), and dispatches the contraction between cuBLAS and a persistent Triton kernel depending on the input shape. The resulting design preserves a strong hybrid backbone while refining individual stages through more specialized execution paths.

\definecolor{solutionbg}{HTML}{F7F8FA}
\definecolor{solutionframe}{HTML}{D7DDE5}
\definecolor{solutionkeyword}{HTML}{2F5877}
\definecolor{solutioncomment}{HTML}{66717F}
\definecolor{solutionstring}{HTML}{76536B}
\definecolor{solutionnumber}{HTML}{8993A0}

\lstdefinestyle{solutioncode}{
  basicstyle=\ttfamily\fontsize{7.2pt}{8.25pt}\selectfont,
  keywordstyle=\color{solutionkeyword}\bfseries,
  commentstyle=\color{solutioncomment}\itshape,
  stringstyle=\color{solutionstring},
  numbers=left,
  numberstyle=\color{solutionnumber}\fontsize{6.2pt}{7pt}\selectfont,
  numbersep=8pt,
  stepnumber=1,
  frame=single,
  framerule=0.45pt,
  rulecolor=\color{solutionframe},
  backgroundcolor=\color{solutionbg},
  breaklines=true,
  breakatwhitespace=false,
  columns=fullflexible,
  keepspaces=true,
  showstringspaces=false,
  emptylines=0,
  tabsize=2,
  captionpos=t,
  aboveskip=8pt,
  belowskip=7pt,
  xleftmargin=1.8em,
  framexleftmargin=1.35em,
  framesep=5pt
}

\tcolorboxenvironment{lstlisting}{
  enhanced jigsaw, breakable, blanker,
  title after break={Listing \thelstlisting\ (continued)},
  colbacktitle=white, coltitle=solutioncomment,
  fonttitle=\small\itshape,
  pad at break*=2mm,
  before skip=0pt, after skip=0pt
}

\subsection{Polyomino Packing}

The solver that scores 91.8907 evaluates every valid orientation and horizontal
position against the current skyline. Its placement key combines the resulting
height, wasted space, boundary slope, and piece area. Simulated annealing then
changes the piece priorities and board width while retaining the best complete
packing. Listing~\ref{lst:polyomino-solution} shows the construction and
refinement loop.

\nolinenumbers
\begin{lstlisting}[
  style=solutioncode,
  language=C++,
  caption={Selected implementation of skyline construction and second-stage refinement.},
  label={lst:polyomino-solution}
]
static bool packForWidthDynamic(
    int W, const vector<Piece>& pieces,
    vector<array<int, 4>>& place, int& outH,
    double timeLimit, const vector<long long>& bias,
    vector<int>* outOrder, int fragMode) {
  const int n = pieces.size();
  vector<int> skyline(W, 0);
  vector<bool> used(n, false);
  int maxH = 0;

  for (int step = 0; step < n; ++step) {
    if ((step & 3) == 0 && elapsed() > timeLimit)
      return false;

    array<long long, 5> globalBest;
    fill(globalBest.begin(), globalBest.end(), LLONG_MAX);
    int bestPiece = -1, bestOrient = -1;
    int bestX = -1, bestY = -1;

    for (int pi = 0; pi < n; ++pi) {
      if (used[pi]) continue;

      array<long long, 5> pieceBest;
      fill(pieceBest.begin(), pieceBest.end(), LLONG_MAX);
      int pieceOrient = -1, pieceX = -1, pieceY = -1;

      for (int oid = 0; oid < pieces[pi].orients.size(); ++oid) {
        const Orient& orient = pieces[pi].orients[oid];
        if (orient.w > W) continue;

        for (int px = 0; px <= W - orient.w; ++px) {
          int py = 0;
          for (auto [cx, cy] : orient.cells)
            py = max(py, skyline[px + cx] - cy);

          int resultingH = maxH;
          long long waste = 0;
          for (auto [cx, maxCy] : orient.colInfo) {
            int top = py + maxCy + 1;
            resultingH = max(resultingH, top);
            waste += top - skyline[px + cx];
          }

          long long slope = 0;
          int leftTop = py + orient.colInfo.front().second + 1;
          int rightTop = py + orient.colInfo.back().second + 1;
          if (px > 0)
            slope += abs(leftTop - skyline[px - 1]);
          if (px + orient.w < W)
            slope += abs(rightTop - skyline[px + orient.w]);

          array<long long, 5> key = {
              resultingH, waste, slope, py,
              static_cast<long long>(orient.w) * orient.h};
          if (key < pieceBest) {
            pieceBest = key;
            pieceOrient = oid;
            pieceX = px;
            pieceY = py;
          }
        }
      }

      if (pieceOrient < 0) continue;
      array<long long, 5> biasedKey = pieceBest;
      biasedKey[0] += bias.empty() ? 0 : bias[pi];
      if (fragMode == 1)
        biasedKey[0] += pieces[pi].cells.size() * 3;
      else if (fragMode == 2)
        biasedKey[0] -= pieces[pi].cells.size() * 2;

      if (biasedKey < globalBest) {
        globalBest = biasedKey;
        bestPiece = pi;
        bestOrient = pieceOrient;
        bestX = pieceX;
        bestY = pieceY;
      }
    }

    if (bestPiece < 0) return false;
    used[bestPiece] = true;
    const Orient& orient = pieces[bestPiece].orients[bestOrient];
    for (auto [cx, cy] : orient.cells) {
      int column = bestX + cx;
      int top = bestY + cy + 1;
      skyline[column] = max(skyline[column], top);
      maxH = max(maxH, top);
    }
    place[bestPiece] = {
        bestX - orient.minXf, bestY - orient.minYf,
        orient.R, orient.F};
    if (outOrder) outOrder->push_back(bestPiece);
  }
  outH = maxH;
  return true;
}

vector<long long> currentBias(n, 0);
int currentW = bestWForRefine;
int currentH = bestH;
long long currentArea = bestArea;
vector<array<int, 4>> currentPlace = bestPlace;
vector<int> currentOrder;

while (elapsed() < annealingDeadline) {
  vector<long long> proposedBias = currentBias;
  int proposedW = proposeWidth(
      currentW, bestW, maxPieceWidth, rng);
  perturbPriorities(
      proposedBias, currentOrder, conflictScore, rng);

  vector<array<int, 4>> proposedPlace(n);
  vector<int> proposedOrder;
  int proposedH;
  if (!packForWidthDynamic(
          proposedW, pieces, proposedPlace, proposedH,
          annealingDeadline, proposedBias, &proposedOrder,
          fragmentationMode))
    continue;

  long long proposedArea =
      static_cast<long long>(proposedW) * proposedH;
  double progress = (elapsed() - annealingStart) /
                    (annealingDeadline - annealingStart);
  double temperature =
      startTemperature * pow(endTemperature / startTemperature,
                             min(1.0, progress));

  bool accept = proposedArea <= currentArea;
  if (!accept) {
    double normalizedIncrease =
        static_cast<double>(proposedArea - currentArea) /
        max(1LL, totalCells);
    double probability =
        exp(-normalizedIncrease / max(temperature, 1e-6));
    accept = uniform01(rng) < probability;
  }
  if (!accept) continue;

  currentBias = proposedBias;
  currentW = proposedW;
  currentH = proposedH;
  currentArea = proposedArea;
  currentPlace = proposedPlace;
  currentOrder = proposedOrder;

  if (proposedArea < bestArea ||
      (proposedArea == bestArea && proposedH < bestH)) {
    bestArea = proposedArea;
    bestW = proposedW;
    bestH = proposedH;
    bestPlace = proposedPlace;
  }
}
\end{lstlisting}
\linenumbers

\Needspace{12\baselineskip}
\subsection{TriMul GPU Kernel}

The selected TriMul kernel records a search-time latency of
\(1110.85\,\mu\mathrm{s}\) and a controlled three-run mean of
\(1128.91\,\mu\mathrm{s}\). For \(C=128\), one Triton kernel fuses layer
normalization, five projections, gating, masking, and layout conversion. The
contraction uses cuBLAS for shapes divisible by 128 and a persistent Triton
kernel with progressively smaller tiles for the remaining shapes.
Listing~\ref{lst:trimul-solution} shows the fused input path and dispatch.

\nolinenumbers
\begin{lstlisting}[
  style=solutioncode,
  language=Python,
  morekeywords={triton,jit,constexpr},
  caption={Selected implementation of the fused front end and shape-dependent contraction.},
  label={lst:trimul-solution}
]
@triton.jit
def fused_ln_proj_gate_transpose_kernel(
    x_ptr, norm_w_ptr, norm_b_ptr, proj_w_t_ptr,
    mask_ptr, left_out_ptr, right_out_ptr, gate_out_ptr,
    total_rows, N,
    C: tl.constexpr, H: tl.constexpr, FIVE_H: tl.constexpr,
    BLOCK_ROW: tl.constexpr, BLOCK_C: tl.constexpr,
    HAS_MASK: tl.constexpr,
):
    rows = tl.program_id(0) * BLOCK_ROW + tl.arange(0, BLOCK_ROW)
    row_mask = rows < total_rows
    offs_c = tl.arange(0, BLOCK_C)
    c_mask = offs_c < C
    offs_h = tl.arange(0, H)

    x = tl.load(
        x_ptr + rows[:, None] * C + offs_c[None, :],
        mask=row_mask[:, None] & c_mask[None, :], other=0.0)
    mean = tl.sum(x, axis=1) / C
    centered = tl.where(c_mask[None, :], x - mean[:, None], 0.0)
    variance = tl.sum(centered * centered, axis=1) / C
    normalized = centered * tl.rsqrt(variance + 1e-5)
    weight = tl.load(norm_w_ptr + offs_c, mask=c_mask, other=0.0)
    bias = tl.load(norm_b_ptr + offs_c, mask=c_mask, other=0.0)
    normalized = (normalized * weight[None, :] +
                  bias[None, :]).to(tl.float16)

    if HAS_MASK:
        input_mask = tl.load(mask_ptr + rows, mask=row_mask, other=0.0)

    batch = rows // (N * N)
    row_in_batch = rows % (N * N)
    transposed_base = batch * H * N * N + row_in_batch
    transposed_ptrs = (
        transposed_base[None, :] + offs_h[:, None] * N * N)

    left_w = tl.load(
        proj_w_t_ptr + offs_c[:, None] * FIVE_H + offs_h[None, :],
        mask=c_mask[:, None], other=0.0)
    left = tl.dot(normalized, left_w)
    left_gate_w = tl.load(
        proj_w_t_ptr + offs_c[:, None] * FIVE_H +
        (2 * H + offs_h)[None, :],
        mask=c_mask[:, None], other=0.0)
    left_gate = tl.sigmoid(tl.dot(normalized, left_gate_w))
    left = (left * left_gate).to(tl.float16)
    if HAS_MASK:
        left *= input_mask[:, None].to(tl.float16)
    tl.store(
        left_out_ptr + transposed_ptrs, left.trans(),
        mask=row_mask[None, :])

    right_w = tl.load(
        proj_w_t_ptr + offs_c[:, None] * FIVE_H +
        (H + offs_h)[None, :],
        mask=c_mask[:, None], other=0.0)
    right = tl.dot(normalized, right_w)
    right_gate_w = tl.load(
        proj_w_t_ptr + offs_c[:, None] * FIVE_H +
        (3 * H + offs_h)[None, :],
        mask=c_mask[:, None], other=0.0)
    right_gate = tl.sigmoid(tl.dot(normalized, right_gate_w))
    right = (right * right_gate).to(tl.float16)
    if HAS_MASK:
        right *= input_mask[:, None].to(tl.float16)
    tl.store(
        right_out_ptr + transposed_ptrs, right.trans(),
        mask=row_mask[None, :])

    output_gate_w = tl.load(
        proj_w_t_ptr + offs_c[:, None] * FIVE_H +
        (4 * H + offs_h)[None, :],
        mask=c_mask[:, None], other=0.0)
    output_gate = tl.dot(normalized, output_gate_w)
    tl.store(
        gate_out_ptr + rows[:, None] * H + offs_h[None, :],
        output_gate.to(tl.float16), mask=row_mask[:, None])


def bmm_contraction(left, right, B, H, N, device):
    left = left.view(B * H, N, N)
    right = right.view(B * H, N, N)
    if N % 128 == 0:
        return torch.bmm(left, right.transpose(1, 2))

    if N % 64 == 0:
        block_m, block_n, block_k, num_warps = 64, 64, 64, 4
    elif N % 32 == 0:
        block_m, block_n, block_k, num_warps = 32, 32, 32, 4
    else:
        block_m, block_n, block_k, num_warps = 16, 16, 16, 2

    output = torch.empty(
        (B * H, N, N), dtype=torch.float16, device=device)
    num_sms = torch.cuda.get_device_properties(0).multi_processor_count
    persistent_bmm_kernel[(num_sms,)](
        left, right, output, B * H, N,
        BLOCK_M=block_m, BLOCK_N=block_n, BLOCK_K=block_k,
        GROUP_SIZE_M=4, NUM_SMS=num_sms,
        num_warps=num_warps, num_stages=3)
    return output


def custom_kernel(data):
    x, mask, weights, config = data
    B, N, _, C = x.shape
    H = config["hidden_dim"]
    total_rows = B * N * N

    if C == 128:
        projection_weights = torch.cat([
            weights["left_proj.weight"],
            weights["right_proj.weight"],
            weights["left_gate.weight"],
            weights["right_gate.weight"],
            weights["out_gate.weight"],
        ]).to(torch.float16).t().contiguous()

        left = torch.empty(
            (B, H, N, N), dtype=torch.float16, device=x.device)
        right = torch.empty_like(left)
        output_gate = torch.empty(
            (total_rows, H), dtype=torch.float16, device=x.device)

        block_rows = 64
        grid = (triton.cdiv(total_rows, block_rows),)
        fused_ln_proj_gate_transpose_kernel[grid](
            x.reshape(total_rows, C),
            weights["norm.weight"], weights["norm.bias"],
            projection_weights,
            mask.reshape(-1) if mask is not None else None,
            left, right, output_gate, total_rows, N,
            C=C, H=H, FIVE_H=5 * H,
            BLOCK_ROW=block_rows, BLOCK_C=C,
            HAS_MASK=(mask is not None),
            num_warps=4, num_stages=3)
    else:
        normalized = layernorm_to_fp16(x, weights)
        projected = F.linear(
            normalized, combined_projection_weights(weights))
        left, right, output_gate = gate_and_transpose(
            projected, mask, B, H, N)

    contracted = bmm_contraction(left, right, B, H, N, x.device)
    return fused_output(contracted, output_gate, weights)
\end{lstlisting}
\linenumbers

\Needspace{14\baselineskip}
\subsection{Lasso Regularization Path}
\label{app:lasso-code}

The 8B-guidance/GLM solution achieves a frozen in-domain score of
\(0.1739 \pm 0.0015\). Its C++ core follows a Lasso homotopy path:
it updates the active-set Cholesky factor as features enter, screens inactive
features using a segment tree of cached correlations, and interpolates between
path knots for the requested regularization parameters.
Listing~\ref{lst:lasso-solution} shows these three components.
The Python wrapper, SIMD primitives, feature-drop updates, and path-state
bookkeeping are omitted; the numbered blocks are excerpts from different parts
of the same solver.

\nolinenumbers
\begin{lstlisting}[
  style=solutioncode,
  language=C++,
  caption={Selected Lasso code: active-set updates, screening, and interpolation.},
  label={lst:lasso-solution}
]
// 1. Incremental factorization for an entering feature.
auto lsolve = [&](double* x, const double* b) {
    for (int i = 0; i < Lk; i++) {
        const double* Li = &Lbuf[i * maxK];
        double s = b[i] - dot_simd(Li, x, i);
        x[i] = s / Li[i];
    }
};

auto usolve = [&](double* x, double* b) {
    for (int j = Lk - 1; j >= 0; j--) {
        x[j] = b[j] / Lbuf[j * maxK + j];
        double xj = x[j];
        if (j > 0) axpy_neg(b, &Lbuf[j * maxK], xj, j);
    }
};

auto chol_update = [&](int j_new) -> bool {
    if (useGram) {
        for (int i = 0; i < nActive; i++)
            vbuf[i] = Gram(activeIdx[i], j_new);
    } else {
        const double* xj_new = &X.data()[(size_t)j_new * n];
        #pragma omp parallel for schedule(static) if(nActive > 32)
        for (int i = 0; i < nActive; i++)
            vbuf[i] = dot_simd(active_cols[i], xj_new, n) / dn;
    }
    lsolve(zbuf.data(), vbuf.data());
    double dd = Gdiag(j_new) - dot_simd(zbuf.data(), zbuf.data(), Lk);
    if (dd <= 1e-20) return false;
    std::memcpy(&Lbuf[Lk * maxK], zbuf.data(), Lk * sizeof(double));
    Lbuf[Lk * maxK + Lk] = std::sqrt(dd);
    Lk++;
    if (useGram) {
        std::memcpy(&G_active[(size_t)nActive * p],
                    &Gram.data()[(size_t)j_new * p],
                    (size_t)p * sizeof(double));
    }
    return true;
};

// 2. Inner path-search step: lazy correlation screening.
// Direction computation and feature-drop handling are omitted.
// Segment tree-based gamma_hat screening
double gamma_best = std::isfinite(gamma_lasso) ? gamma_lasso : INF;
int j_new = -1; double s_new = 0.0;

checked.clear();

while (true) {
    // O(1) max query: read root (branchless tree)
    double stale_cj = segVal[1];
    int j = segIdx[1];

    if (j < 0) break; // no inactive features

    if (activeMaskD[j] > 0.5) {
        segDeactivate(j);
        continue;
    }

    // Safe screening: O(1) check, no push-back needed
    double safe_thresh = C - 2.0 * A_A * gamma_best - delta_accum;
    if (std::isfinite(gamma_best) && safe_thresh > 0 && stale_cj < safe_thresh) {
        break; // all remaining features have smaller stale|c_j|
    }

    // Temporarily remove from tree for consideration
    segDeactivate(j);

    // Compute actual c_j and a_j (lazy evaluation)
    double cj, aj;
    compute_cj_aj(j, cj, aj);

    double abs_cj = std::fabs(cj);

    // Exact screening
    double exact_thresh = C - 2.0 * A_A * gamma_best;
    if (std::isfinite(gamma_best) && exact_thresh > 0 && abs_cj < exact_thresh) {
        checked.emplace_back(abs_cj, j);
        continue;
    }

    // Compute gamma_j
    double d1 = A_A - aj, d2 = A_A + aj;
    double gp = (d1 > eps) ? (C - cj) / d1 : INF;
    double gm = (d2 > eps) ? (C + cj) / d2 : INF;
    if (gp < 0) gp = INF;
    if (gm < 0) gm = INF;

    if (gp < gamma_best) { gamma_best = gp; j_new = j; s_new = 1.0; }
    if (gm < gamma_best) { gamma_best = gm; j_new = j; s_new = -1.0; }

    checked.emplace_back(abs_cj, j);
}

// Restore checked features to tree with updated |c_j|
for (auto& pr : checked) {
    segUpdate(pr.second, pr.first);
}

double gamma_hat = (j_new >= 0) ? gamma_best : INF;
double gamma = std::min(gamma_hat, gamma_lasso);

// 3. Interpolate the requested regularization parameters.
int nk = (int)knot_lambda.size();
VectorXd coef_path = VectorXd::Zero(p * n_lambda);

if (nk >= 1) {
    VectorXd b(p);
    int seg = 0;
    for (int kk = 0; kk < n_lambda; kk++) {
        double lam = lambda_path(kk);
        b.setZero();

        if (lam >= knot_lambda[0] - 1e-15) {
            continue;
        }

        if (lam <= knot_lambda[nk-1] + 1e-15) {
            for (auto& kv : knot_sparse[nk-1])
                b(kv.first) = kv.second;
            coef_path.segment(kk * p, p) = b;
            continue;
        }

        int lo = 0, hi = nk - 1;
        while (lo < hi - 1) {
            int mid = (lo + hi) / 2;
            if (knot_lambda[mid] > lam) lo = mid;
            else hi = mid;
        }
        seg = lo;

        double la = knot_lambda[seg], lb = knot_lambda[seg+1];
        if (std::fabs(la - lb) < 1e-15) {
            for (auto& kv : knot_sparse[seg])
                b(kv.first) = kv.second;
        } else {
            double t = (la - lam) / (la - lb);
            for (auto& kv : knot_sparse[seg])
                b(kv.first) = (1.0 - t) * kv.second;
            for (auto& kv : knot_sparse[seg+1])
                b(kv.first) += t * kv.second;
        }

        for (auto& kv : knot_sparse[seg])
            if (std::fabs(b(kv.first)) < 1e-12) b(kv.first) = 0.0;
        for (auto& kv : knot_sparse[seg+1])
            if (std::fabs(b(kv.first)) < 1e-12) b(kv.first) = 0.0;

        coef_path.segment(kk * p, p) = b;
    }
}
\end{lstlisting}
\linenumbers

\Needspace{14\baselineskip}
\subsection{AHC058: Apple Production Planning}
\label{app:ahc058-code}

The 14B-guidance/GLM solution scores \(850{,}082{,}731\) on AtCoder.
Its constructive search scores machine upgrades by their cost and remaining
production horizon, including interactions across production levels.
A two-stage sigmoid schedule adjusts the scoring blend, using the current
machine populations and powers during the transition.
Listing~\ref{lst:ahc058-solution} shows the schedule and the upgrade score.
Short lookahead, state updates, auxiliary scoring functions, and the outer
search and refinement stages are omitted.

\nolinenumbers
\begin{lstlisting}[
  style=solutioncode,
  language=C++,
  caption={Selected AHC058 code: phase scheduling and state-dependent upgrade scores.},
  label={lst:ahc058-solution}
]
// 1. Adapt the scoring blend to time and production state.
// Dual-sigmoid blend with real-time state adaptation
void set_synergy_blend_dual(int turn, int total, const State& s) {
    double progress = (double)turn / (double)total;

    // First sigmoid: early phase transition around 15%
    double k1 = 12.0;
    double mid1 = 0.15;
    double sig1 = 1.0 / (1.0 + exp(-k1 * (progress - mid1)));

    // Second sigmoid: late phase transition around 65%
    double k2 = 12.0;
    double mid2 = 0.65;
    double sig2 = 1.0 / (1.0 + exp(-k2 * (progress - mid2)));

    if (progress < 0.30) {
        // Early "population growth" phase: favor single-level efficiency
        g_synergy_blend = 0.05 + 0.10 * sig1;
    } else if (progress > 0.60) {
        // Late "cascading synergy" phase: emphasize multi-level interactions
        g_synergy_blend = 0.30 + 0.70 * sig2;
    } else {
        // Transition phase (30-60%): dynamically adjust based on B ratios across levels
        double b_low = 0, b_high = 0;
        for (int j = 0; j < N; j++) {
            b_low += s.B[0][j] + s.B[1][j];
            b_high += s.B[2][j] + s.B[3][j];
        }
        double b_ratio = b_high / max(b_low, 1.0);
        double maturity = min(1.0, b_ratio);

        // Also consider power distribution across levels
        double p_low = 0, p_high = 0;
        for (int j = 0; j < N; j++) {
            p_low += (double)s.P[0][j] + (double)s.P[1][j];
            p_high += (double)s.P[2][j] + (double)s.P[3][j];
        }
        double p_ratio = p_high / max(p_low + p_high, 1.0);
        double power_maturity = min(1.0, p_ratio * 2.0);

        double dynamic_weight = 0.5 * maturity + 0.5 * power_maturity;

        double early_val = 0.05 + 0.10 * sig1;
        double late_val = 0.30 + 0.70 * sig2;
        g_synergy_blend = early_val + dynamic_weight * (late_val - early_val);
    }

    g_synergy_blend = max(0.0, min(1.0, g_synergy_blend));
}

// 2. Score an affordable upgrade over the remaining horizon.
double compute_benefit(const State& s, int i, int j, int R,
                       const vector<double>& mults, double proj_factor) {
    if (R <= i) return -1e18;
    double cost = (double)C[i][j] * (double)(s.P[i][j] + 1);
    if (cost > s.apples) return -1e18;
    double benefit = (double)A[j] * s.B[i][j];
    double cascade_product = 1.0;
    for (int k = 0; k < i; k++) {
        double pk = (double)s.P[k][j];
        if (pk < 1) pk = proj_factor;
        cascade_product *= pk;
    }
    benefit *= cascade_product;
    double compounding = 1.0;
    for (int k = 1; k <= i; k++) {
        double pk = (double)s.P[k][j];
        if (pk < 1) pk = proj_factor;
        double growth_ratio = s.B[k][j] * pk / max(s.B[k-1][j], 1.0);
        compounding *= (1.0 + min(growth_ratio, 10.0) * 0.0005 * (double)min(R, 200));
    }
    for (int k = 0; k < i; k++) {
        double pk = (double)s.P[k][j];
        if (pk < 1) pk = proj_factor;
        compounding *= (1.0 + min(pk, 1000.0) * 0.001 * (double)min(R, 100) / 100.0);
    }
    benefit *= compounding;
    if (i >= 1 && i < L - 1) {
        double pk_i = max((double)s.P[i][j], proj_factor);
        double next_B = s.B[i+1][j];
        double marginal = next_B * pk_i - s.B[i][j];
        if (marginal > 0) {
            double ratio = marginal / max(s.B[i][j], 1.0);
            benefit *= (1.0 + 0.05 * min(log2(1.0 + ratio), 5.0));
        }
    }
    if (i == 0 && L > 1) {
        double p0 = max((double)s.P[0][j], proj_factor);
        double marginal = s.B[1][j] * p0 - s.B[0][j];
        if (marginal > 0) {
            double ratio = marginal / max(s.B[0][j], 1.0);
            benefit *= (1.0 + 0.03 * min(log2(1.0 + ratio), 5.0));
        }
    }
    double discount = pow(0.998, (double)i);
    benefit *= discount;
    double b = binom(R, i + 1);
    benefit *= (b * mults[i]);
    if (benefit <= 0 || cost <= 0) return -1e18;
    double lp_term = 0.0;
    if (g_use_lp) {
        lp_term = log2(g_level_priority[i][j]);
    }
    double base_score = log2(benefit) - log2(cost) + g_weight_log[i][j] + lp_term;
    if (g_use_cascade) {
        set_phase_weights(T - R, T);
        set_discount_rate_state(s, T - R, T);
        double cb = compute_cascading_benefit(s, i, j, R);
        if (cb > -1e17) {
            return 0.85 * base_score + 0.15 * cb;
        }
    }
    // Dual-sigmoid phase blend: early game favors single-level, late game favors multi-level
    if (g_synergy_blend > 0.01 && i > 0) {
        double single_level = log2(max((double)A[j] * s.B[i][j], 1e-18)) - log2(max(cost, 1.0));
        return (1.0 - g_synergy_blend) * single_level + g_synergy_blend * base_score;
    }
    return base_score;
}

\end{lstlisting}
\linenumbers

\end{document}